\documentclass{article}

\usepackage{microtype}
\usepackage{graphicx}
\usepackage{subfigure}
\usepackage{booktabs}

\usepackage{hyperref}
\usepackage{caption}

\usepackage[arxiv]{tech_report}

\usepackage{amsmath}
\usepackage{amssymb}
\usepackage{mathtools}
\usepackage{amsthm}
\usepackage{multirow}

\usepackage[capitalize,noabbrev]{cleveref}

\theoremstyle{plain}

\theoremstyle{definition}

\theoremstyle{remark}

\usepackage[textsize=tiny]{todonotes}

\icmltitlerunning{CHATS: Combining Human-Aligned Optimization and Test-Time Sampling for Text-to-Image Generation}

\begin{document}

\twocolumn[
\icmltitle{CHATS: Combining Human-Aligned Optimization and Test-Time Sampling for Text-to-Image Generation}

\icmlsetsymbol{intern}{$^\dagger$}

\begin{icmlauthorlist}
\icmlauthor{Minghao Fu}{lab,school,inc,intern}
\icmlauthor{Guo-Hua Wang}{inc}
\icmlauthor{Liangfu Cao}{inc}
\icmlauthor{Qing-Guo Chen}{inc}
\icmlauthor{Zhao Xu}{inc}
\icmlauthor{Weihua Luo}{inc}
\icmlauthor{Kaifu Zhang}{inc}
\end{icmlauthorlist}

\icmlaffiliation{school}{School of Artificial Intelligence, Nanjing University}
\icmlaffiliation{lab}{National Key Laboratory for Novel Software Technology, Nanjing University}
\icmlaffiliation{inc}{Alibaba Group}

\icmlcorrespondingauthor{Minghao Fu}{fumh@lamda.nju.edu.cn}
\icmlcorrespondingauthor{Guo-Hua Wang}{wangguohua@alibaba-inc.com}

{
\vspace{0.3em}
\includegraphics[width=0.99\textwidth]{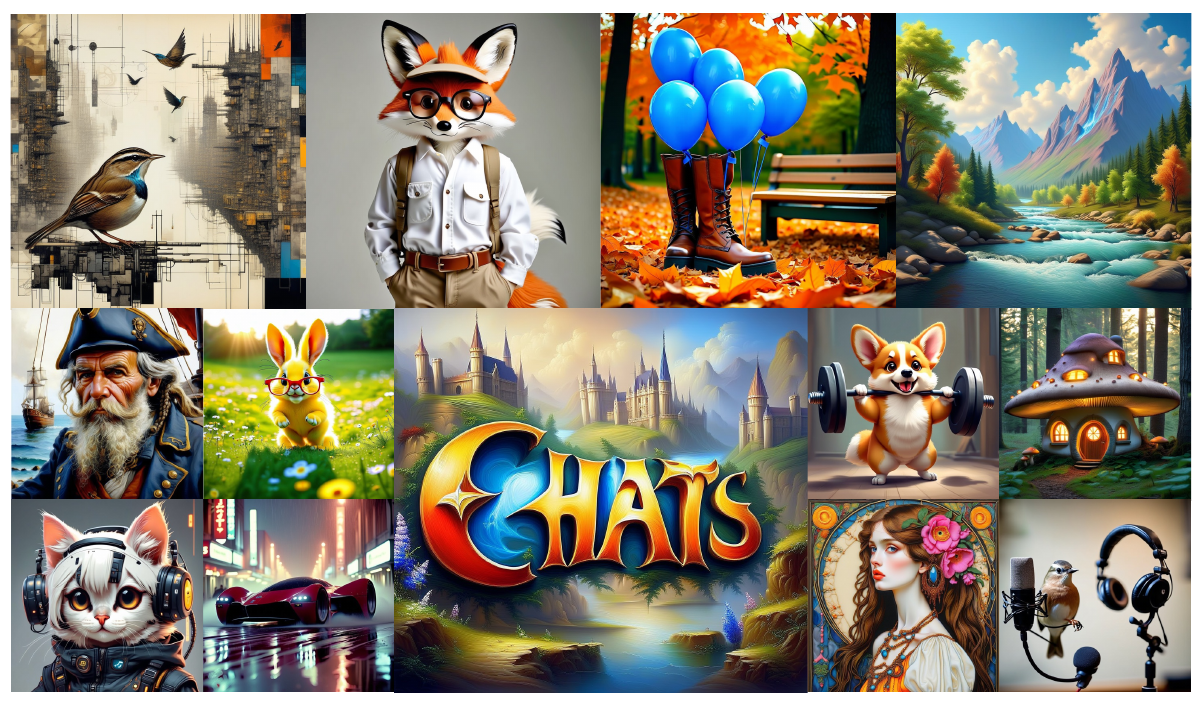}%
\vspace{-1em}
\captionsetup[figure]{hypcap=false}
\captionof{figure}{
Images generated by CHATS with SDXL, an innovative text-to-image generation framework that, for the first time, facilitates collaboration between human preference alignment and test-time sampling, establishing a new paradigm in text-conditioned image synthesis.
\vspace{-1.5em}
}
\label{fig:general_illustration}%
}

\vskip 0.3in
]

\printAffiliationsAndNotice{$^\dagger$ Work done during the internship at Alibaba Group.}

\begin{abstract}
Diffusion models have emerged as a dominant approach for text-to-image generation. Key components such as the human preference alignment and classifier-free guidance play a crucial role in ensuring generation quality. However, their independent application in current text-to-image models continues to face significant challenges in achieving strong text-image alignment, high generation quality, and consistency with human aesthetic standards. In this work, we for the first time, explore facilitating the collaboration of human performance alignment and test-time sampling to unlock the potential of text-to-image models. Consequently, we introduce \textbf{CHATS} (\textbf{C}ombining \textbf{H}uman-\textbf{A}ligned optimization and \textbf{T}est-time \textbf{S}ampling), a novel generative framework that separately models the preferred and dispreferred distributions and employs a proxy-prompt-based sampling strategy to utilize the useful information contained in both distributions. We observe that CHATS exhibits exceptional data efficiency, achieving strong performance with only a small, high-quality funetuning dataset. Extensive experiments demonstrate that CHATS surpasses traditional preference alignment methods, setting new state-of-the-art across various standard benchmarks. The code is publicly available at \href{https://github.com/AIDC-AI/CHATS}{github.com/AIDC-AI/CHATS}.
\end{abstract}

\section{Introduction}

\begin{figure*}
    \centering
    \includegraphics[width=.8\linewidth]{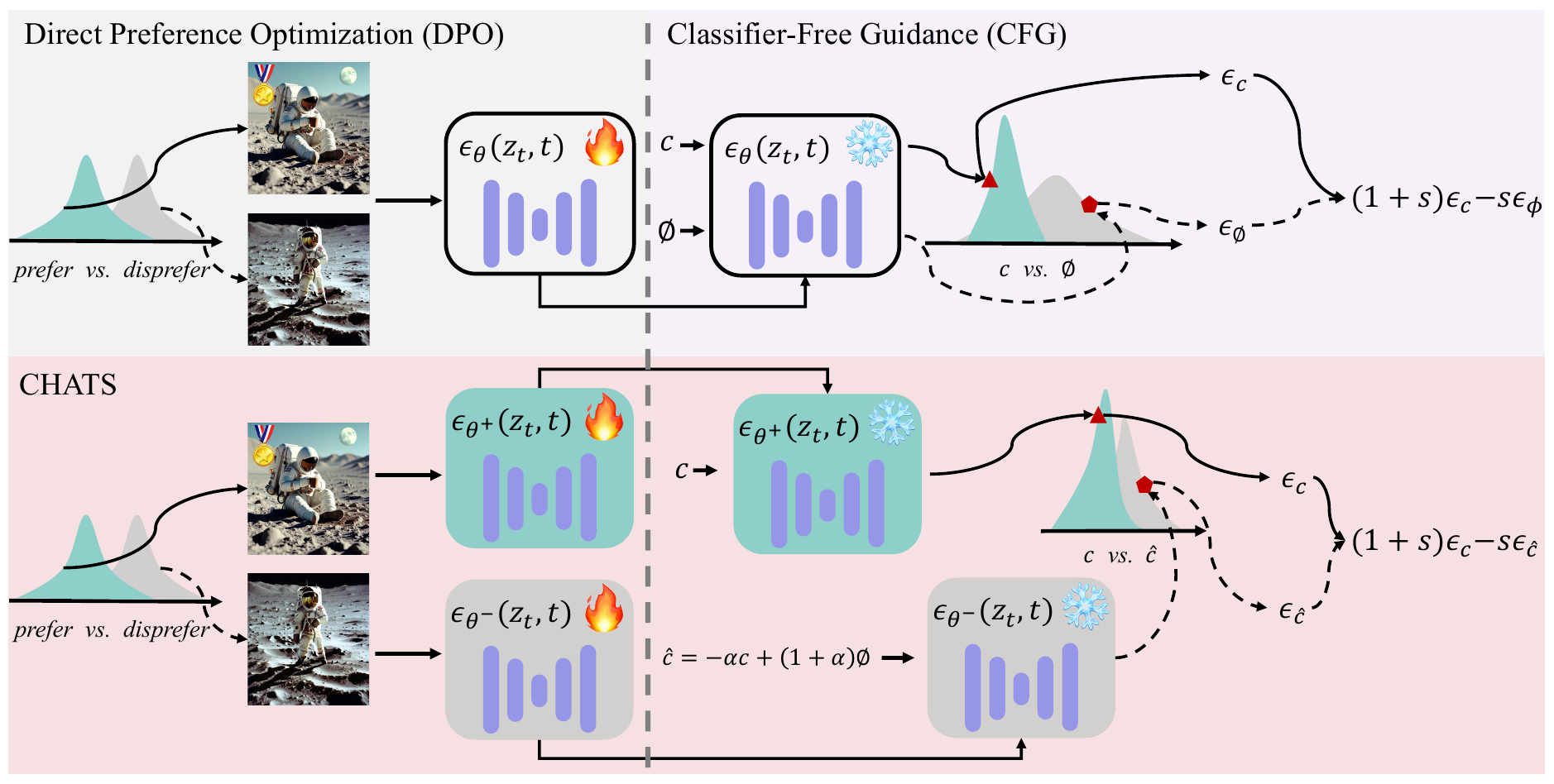}
    \caption{Illustration of CHATS. Both DPO and CFG adopt preference-based frameworks in their formulations. DPO derives preferences from user feedback, while CFG utilizes diverse prompting strategies to capture such discrepancies. However, their current independent working paradigms present significant potential for integration. Our CHATS addresses this by separately modeling preferred and dispreferred information with two distinct models. Additionally, it employs a proxy-prompt-based sampling strategy to facilitate effective collaboration between two models.}
    \label{fig:network}
\end{figure*}

Diffusion models~\cite{diffusion_survey} have emerged as a leading generative framework~\cite{vae,diffusion,gan} by iteratively transforming random noise into structured images~\cite{ddpm,ldm,diffusion_sde}. Text-to-image generation, as a key application, offers broad versatility, supporting tasks like image editing~\cite{instructpix2pix} and generative conversational AI~\cite{gpt3}.

In particular, high-quality image generation largely depends on two stages: 1) human preference alignment, in which ranked pairs reflecting human preference are utilized to train the model for maximizing the distinction between preferred and dispreferred generated images; and 2) sampling, where Classifier-Free Guidance (CFG)~\cite{cfg} plays a vital role by merging conditional and unconditional noise predictions to focus on regions of higher text-conditioned density.

Preference alignment involves human experts categorizing images into preferred and dispreferred groups based on their aesthetic desirability, and using these two groups of images to train a model for capturing desirable characteristics from preferred images while avoiding undesirable features or errors present in dispreferred ones. Similarly, the sampling procedure of CFG can be conceptualized as denoising from a \textit{preferred} distribution (conditional) while simultaneously steering away from a \textit{dispreferred} distribution (unconditional). This dual-framework naturally aligns with the principles of preference alignment. Nevertheless, previous efforts~\cite{cg,DPOK,d3po,pag,ddpo} focus on optimizing these two processes in isolation, leaving the integration of the two for leveraging their mutual benefits unexplored and resulting in suboptimal outcomes. This presents an opportunity for us to develop unified method that enables a more effective solution.

Motivated by this, we introduce \textbf{CHATS} (\textbf{C}ombining \textbf{H}uman-\textbf{A}ligned optimization and \textbf{T}est-time \textbf{S}ampling), a novel approach that integrates human preference alignment with the sampling stage, thereby further enhancing the performance of text-to-image generation systems.

Specifically, CHATS employs two distinct models: one tasked with learning the preferred distribution (referred to as the \textit{preferred model}) and another dedicated to capturing the dispreferred and unconditional distributions (referred to as the \textit{dispreferred model}). These models are trained using a novel, theoretically grounded objective function based on the general principle of direct preference optimization (DPO)~\cite{llm_dpo}. During sampling, two models collaborate to guide the denoising process towards the preferred model while concurrently repelling it from the dispreferred model using a proxy-prompt-based sampling strategy to leverage conditional and unconditional information.

Our method is simple, intuitive, has a robust theoretical foundation, while also demonstrating high efficiency and effectiveness. In particular, training our approach requires only 7,459 high-quality preference pairs, exhibiting significantly greater data efficiency compared to traditional preference optimization methods~\cite{diffusion-dpo}.

Empirical evaluations on two mainstream text-to-image generation frameworks, diffusion~\cite{sdxl} and flow matching~\cite{rectified_flow,flow_matching}, underscore the superiority of our method. Our approach not only surpasses naive CFG sampling strategies but also outperforms the most effective preference optimization methods currently available~\cite{diffusion-dpo}. Furthermore, the default configuration of hyperparameters in our method performs robustly across different model architectures, highlighting its practical utility.

As far as we know, our work marks the first attempt to unify the human preference alignment and sampling stages within the current text-to-image generation framework. We hope that our research can shed new light on the paradigm design of such generative systems.

The contributions of this paper can be summarized as:
\begin{itemize}
    \item We are the first to integrate human preference optimization finetuning with the sampling process, leveraging their inherent synergy to refine image generation.
    \item We propose CHATS, a novel framework that offers precise control over preference distributions while ensuring efficient sampling. CHATS requires only a small high-quality dataset for training, maintaining high data efficiency.
    \item Extensive experiments are conducted to demonstrate that the collaboration between preference alignment and sampling together yields significantly improved generation performance compared to baseline methods.
\end{itemize}

\section{Related Work}

\subsection{Human Preference Optimization}

Human preference optimization aims to improve the aesthetic quality of generated images. One line of research~\cite{image_reward,hpsv2,pickscore,ahf} focused on developing reward models to evaluate the alignment between generated images and their corresponding prompts. These reward models are subsequently used to finetune diffusion models, guiding them to better capture desirable characteristics. 

Another line of research explored direct finetuning on preferred images~\cite{raft,emu} or adopted reinforcement learning from human feedback (RLHF)~\cite{ddpo,DPOK,d3po,diffusion-dpo}. Specifically, DDPO~\cite{ddpo} formulated denoising as a multi-step decision-making problem. DPOK~\cite{DPOK} incorporated KL regularization as an implicit reward to stabilize the finetuning process. D3PO~\cite{d3po} and Diffusion-DPO~\cite{diffusion-dpo} extended the concept of DPO~\cite{llm_dpo} to diffusion models by optimizing the respective policies. CHATS departs from these approaches by jointly modeling both preferred and dispreferred distributions with two coupled networks. 

Some more recent methods include NPO~\cite{npo}, which employs an unlearning based framework to incorporate negative preference signals, and Diffusion-NPO~\cite{diffusion_npo}, which trains a negative preference model to substitute the unconditional branch at sampling rather than jointly modeling both distributions. See Section~\ref{supp:more_related_works} for further details.

\subsection{Guided Sampling for Text-to-Image Generation}

Guided sampling enhances fidelity by steering generation towards specific directions determined by various strategies. CG~\cite{cg} utilized the gradients from a classifier as the signal to achieve class-conditional generation at the cost of diversity. CFG~\cite{cfg} extended this concept to the open-vocabulary domain by combining conditional and unconditional outputs, thereby biasing the noise towards areas with higher semantic density. Numerous methods~\cite{sag,uni_guide,readout,glide} were proposed to further refine CFG. In contrast to these approaches, as shown in Fig.~\ref{fig:network}, our CHATS enhances sampling quality by facilitating collaboration with DPO, enabling both processes to mutually reinforce each other and thus achieve superior results.

\section{Preliminaries}
\subsection{Diffusion Models}
Diffusion-based generative frameworks~\cite{diffusion,ddpm} operate by iteratively injecting Gaussian noise $\epsilon$ sampled from $\mathcal{N}(\mathbf{0}, \mathbf{I})$ into clean data $z_0$, creating progressively noisier versions $z_t$ at each timestep $t$. This forward diffusion is described by:
\begin{equation}
\label{eq:ddpm_to_xt}
q(z_t \mid z_0) = \mathcal{N}\bigl(z_t; \sqrt{\bar{\alpha}_t}\,z_0,\sqrt{1-\bar{\alpha}_t}\,\mathbf{I}\bigr),
\end{equation}
where $\alpha_t = 1-\beta_t$ and $\bar{\alpha}_t = \prod_{s=1}^t \alpha_s$ emerge from a predetermined schedule $\{\beta_1, \dots, \beta_T\}$. The denoising model $\epsilon_\theta$ is trained to predict the noise $\epsilon$, under the objective:
\begin{equation}
\label{eq:ddpm_obj}
\mathcal{L}_{\text{Diffusion}} 
= \mathbb{E}_{z_t, t, \epsilon}
\bigl\|\epsilon - \epsilon_\theta(z_t, t)\bigr\|^2,
\end{equation}
where $t$ is uniformly sampled from $\{1, \dots, T\}$. 

\subsection{Rectified Flow}
Rectified Flow~\cite{rectified_flow,flow_matching} learns a direct mapping from noise to target data through a reversible ODE: 
\begin{equation}
\label{eq:flow_ode}
    \frac{d z_t}{dt} = v_{\theta}(z_t, t).
\end{equation}
The intermediate state $z_t$ during training is approximated as:
\begin{equation}
\label{eq:flow_xt}
    z_t = (1-t) z_0 + t \epsilon,
\end{equation}
where $t \in [0, 1]$ controls the progression from $z_0$ to noise $\epsilon$.

\subsection{Direct Preference Optimization}

DPO~\cite{llm_dpo} formalizes the preference optimization problem as a single-stage policy training based on human preference data, without explicitly relying on a reward model. Follow this rule, given condition $c$, only preference-ranked pairs $z_0^{+} \succ z_0^{-}$ (\textit{preferred vs. non-preferred}) are available. The Bradley-Terry~\cite{BT_model} model characterizes the human preference distribution as:
\begin{equation}
p(z_0^+ \succ z_0^- | c) = \sigma \big(r(z_0^+, c) - r(z_0^-, c)\big),
\end{equation}
where $\sigma(\cdot)$ denotes the sigmoid function and $r$ is a reward model that quantifies the coherence between the input image and text. Approximated by a neural network parameterized by $\theta$, $r$ is optimized by minimizing the negative log-likelihood, as follows:
$$
\mathcal{L}_{\text{DPO}} = - \mathbb{E}_{(c, z_0^+, z_0^-) \sim \mathcal{D}} \left[\log p_\theta(z_0^+ \succ z_0^- | c)\right],
$$
where $\mathcal{D} = \{(c, z_0^+, z_0^-)\}$ represents the preference dataset obtained through human feedback. They build on the work of \citet{sequence_tutor,human_centric_rl} by implicitly employing a reward model to optimize the task distribution $p_{\theta}$ using reference distribution $p_{\text{ref}}$ and demonstrating that $p_{\theta}$ admits a global optimum. Leveraging this property, DPO derives the following final preference alignment objective:
\begin{equation}
\label{eq:dpo_obj}
\textstyle
\mathcal {L}_{\text{DPO}} \!=\! - \mathbb{E}_{(c, z_0^+, z_0^-) \sim \mathcal{D}}\!\! \left[
\log\sigma\!\!\left(\log\! \frac{p_{\theta}(z_0^+| c)}{p_{\text{ref}}(z_0^+|c)} \!-\! \log \!\frac{p_{\theta}(z_0^-|c)}{p_{\text{ref}}(z_0^-|c)}\!\right)\!\right]\!.
\end{equation}

\subsection{Classifier-Free Guidance} 

To enhance the alignment between generated images and textual descriptions, Classifier-Free Guidance (CFG)~\cite{cfg} modifies the sampling distribution as:
\begin{equation}
    \label{eq:cfg_sampling}
    \tilde{p}_{\theta}(z_t|c) \propto {p}_{\theta}(z_t|c)^{1+s} {p}_{\theta}(z_t)^{-s},
\end{equation}
where $c$ denotes the textual input, $p_\theta(z_t|c)$ represents the conditional generative distribution, $p_\theta(z_t)$ is the unconditional distribution and $s$ is a scale scalar. CFG guides the generation by estimating the diffusion score~\cite{diffusion_sde} as:
\begin{equation}
\label{eq:cfg_score}
\Tilde{\epsilon}_{\theta}(z_t, c) = (1+s){\epsilon}_{\theta}(z_t, c) - s{\epsilon}_{\theta}(z_t).
\end{equation}
CFG biases the generation process towards the conditional distribution while reducing reliance on the unconditional counterpart, effectively aligning the generated content with the provided textual prompt.

\section{CHATS: Combining Human-Aligned Optimization and Test-Time Sampling}

DPO and CFG both enhance generation performance by leveraging preferred and dispreferred information, yet they rely on a single model to process and balance these conflicting distributions. In contrast, as shown in Fig.~\ref{fig:network}, our CHATS adopts a more nuanced approach by employing two models, initialized from the same reference model, to independently capture the characteristics of the preferred and dispreferred distributions. During sampling, these models collaborate using a proxy-prompt-based conditional strategy, seamlessly integrating their strengths.

\subsection{Training}

We aim for both the preferred and dispreferred models to maximize their reward scores, defined by $r_{\theta^+}$ and $r_{\theta^-}$, respectively. However, the introduction of two independent models brings a new challenge in designing a valid learning objective. In this scenario, the naive DPO loss defined in Eq.~\ref{eq:dpo_obj} is no longer suitable, as its direct use would lead to the collapse of the dispreferred model. To address this issue, we reformulate the training objective as follows:
\begin{equation}
\label{eq:twin_naive_obj}
    \mathcal{L} \!=\! - \mathbb{E}_{(c, z_0^+, z_0^-) \sim \mathcal{D}} \!\left[ \log \sigma \big(r_{\theta^+}(z_0^+, c) \!+\! r_{\theta^-}(z_0^-, c)\big)\! \right]\!,
\end{equation}
where the sum of reward scores from two models is treated as the target. This formulation enables both models to learn meaningful image distributions, with the preferred model capturing high-quality characteristics and the dispreferred model fitting a less optimal but complementary image distribution.

Training using this objective poses significant challenges in deriving the task distributions $p_{\theta^+}(z_{t-1}^+|z_t^+)$ and $p_{\theta^-}(z_{t-1}^-|z_t^-)$, as we need to account for all potential denoising trajectories from $z_T$ to $z_0$ under conditioning $c$. This complexity arises from the intractability of marginalizing over all intermediate steps in the reverse diffusion process.

To address this challenge, following \citet{diffusion-dpo}, we define $R(z_{0:T}^i, c)$ as the reward score associated with the entire denoising chain from $z_T$ to $z_0$ given the condition $c$. We further express the per-sample reward $r(z_0^i, c)$ as the expectation of this reward score over the intermediate steps $z_{1:T}^i$:
\begin{equation}
\label{eq:r}
    r(z_0^i, c) = \mathbb{E}_{p_{\theta^i}(z_{1:T}^i | z_0^i, c)} \left[ R(z_{0:T}^i, c) \right],
\end{equation}

where $i \in \{+, -\}$ represents the preferred and dispreferred distributions, respectively. We abbreviate $c$ to simplify the notation. By substituting Eq.~\ref{eq:r} into Eq.~\ref{eq:twin_naive_obj}, assuming $z_{1:T}^+ \sim p_{\theta^+}(z_{1:T}^+|z_0^+)$ and $z_{1:T}^- \sim p_{\theta^-}(z_{1:T}^-|z_0^-)$, with $z_0^+$ and $z_0^-$ treated as input variables, the loss for each ranked pair is defined as:
\begin{equation}
\textstyle
\mathcal{L} (z_0^+, z_0^-) \!=\!
-\log \!\sigma\! \biggl(\!
{\mathbb{E}}_{\substack{z_{1:T}^+ \\ z_{1:T}^-}}\!\!
\left[ \log\! \frac{p_{\theta^+}(z_{0:T}^+)}{p_{\text{ref}}(z_{0:T}^+)}\! +\! \log\! \frac{p_{\theta^-}(z_{0:T}^-)}{p_{\text{ref}}(z_{0:T}^-)}\!\right]\!\biggr)\ .
\end{equation}

We approximate $p_{\theta^i}(z_{1:T}^i|z_0^i)$ using $q(z_{1:T}|z_0)$ as the direct estimation of $p_{\theta^i}(z_{1:T}^i|z_0^i)$ is intractable. Leveraging Jensen's inequality, given $(z_0^+,z_0^-) \sim \mathcal D$, $t\sim \mathcal U(0,T)$, $z_t^+\sim q(z_t^+|z_0^+)$ and $z_t^-\sim q(z_t^-|z_0^-)$, we obtain the following upper bound:
{\small
\begin{equation}
\label{eq:chats_target}
\begin{split}
\mathcal{L} (z_0^+, z_0^-) &\leq -\mathbb{E}_{t,z_t^+,z_t^-} \log\sigma \bigg[-T \bigg( \\ 
&\mathbb{D}_{\text{KL}} \big( q(z_{t-1}^+|z_{t}^+,z_{0}^+) \big\| p_{\theta^+}(z_{t-1}^+|z_{t}^+) \big) \\
&- \mathbb{D}_{\text{KL}} \big( q(z_{t-1}^+|z_{t}^+,z_{0}^+) \big\| p_{\text{ref}}(z_{t-1}^+|z_{t}^+) \big) \\
&+ \mathbb{D}_{\text{KL}} \big( q(z_{t-1}^-|z_{t}^-,z_{0}^-) \big\| p_{\theta^-}(z_{t-1}^-|z_{t}^-) \big) \\
&- \mathbb{D}_{\text{KL}} \big( q(z_{t-1}^-|z_{t}^-,z_{0}^-) \big\| p_{\text{ref}}(z_{t-1}^-|z_{t}^-) \big) \bigg) \bigg].
\end{split}
\end{equation}}

For diffusion models, the loss function of CHATS is explicitly formulated as:
{\small
\begin{equation}
\begin{split}
\label{eq:diffusion_chat_obj}
\mathcal{L}_\text{CHATS}^{\text{Diffusion}} &= -\mathbb{E}_{(z_0^+, z_0^-), (z^+_t, z^-_t),t}
\log\sigma \bigg[-T \bigg( \\
&+ \big\|\epsilon^+ \!\!-\! \epsilon_{\theta^+} (z^+_t, t)\big\|^2 \!\!-\! \big\|\epsilon^+ - \epsilon_{\text{ref}} (z^+_t, t)\big\|^2 \\
&+ \big\|\epsilon^- \!\!-\! \epsilon_{\theta^-} (z^-_t, t)\big\|^2 \!\!-\! \big\|\epsilon^- - \epsilon_{\text{ref}} (z^-_t, t)\big\|^2 \!\bigg) \!\bigg]\!.
\end{split}
\end{equation}}

For flow matching models, $q(z_{t-1}|z_t, z_0)$ is expressed using the Dirac delta function. In this setting, $q$ is uniquely determined by Eq.~\ref{eq:flow_ode}. Consequently, the loss function is defined as:
{\small
\begin{equation}
\begin{split}
\label{eq:flow_chat_obj}
\mathcal{L}_\text{CHATS}^{\text{Flow}} &= -\mathbb{E}_{(z_0^+, z_0^-), (z^+_t, z^-_t), t}\log\sigma \bigg[-T \bigg( \\
& + \big\|v^+ \!\!-\! v_{\theta^+} (z^+_t, t)\big\|^2 \!\!-\! \big\|v^+ - v_{\text{ref}} (z^+_t, t)\big\|^2 \\
& + \big\| v^- \!\!-\! v_{\theta^-} (z^-_t, t)\big\|^2 \!\!-\! \big\| v^- - v_{\text{ref}} (z^-_t, t)\big\|^2 \bigg) \bigg].
\end{split}
\end{equation}}

Please refer to Supp.~\ref{sec:supp_training_obj_of_CHATS} for a detailed derivation of the training objective. Our method ensures that the two models, responsible for learning the preferred and dispreferred distributions, achieve the dual objectives of accurately fitting their respective targets while not shifting too far from the reference model. 

\subsection{Sampling}

After training, we obtain two models: one corresponds to the \textit{preferred} distribution, denoted as $p_{\theta^+}$, and the other corresponds to the \textit{dispreferred} distribution, denoted as $p_{\theta^-}$. Motivated by the mechanism of CFG~\cite{cfg}, we propose a novel strategy to leverage the contribution of these two models. Let $z_t$ be the noisy image at step $t$ and $c$ be the condition. The sampling distribution of CHATS is expressed as:
\begin{equation}
\label{eq:twin_sampling}
\textstyle
\tilde{p}_{\theta}(z_t | c) \!\propto\! p_{\theta^+}(z_t | c)^{1+s} p_{\theta^-}\!(z_t | c)^{\alpha s} p_{\theta^-}\!(z_t)^{-(1 + \alpha)s}\!.
\end{equation}

Please refer to Sec~\ref{sec:chats_sampler} for the details of its theoretical foundations. In this formulation, $p_{\theta^+}(z_t | c)^{1+s}$ encourages the samples to align closely with the preferred distribution. When $0 < \alpha < \frac{1+s}{s}$, the dispreferred distribution $p_{\theta^-}$ is partially incorporated, allowing it to contribute potentially useful patterns while remaining less influential than $p_{\theta^+}$. Conversely, when $\alpha < 0$, the terms $p_{\theta^-}(z_t | c)^{\alpha s}$ and $p_{\theta^-}(z_t)^{-(1+\alpha)s}$ actively push the samples away from undesired modes in the dispreferred distribution, both conditionally and unconditionally.

This balanced approach ensures that the sampling process leverages the strengths of the preferred distribution while selectively integrating or excluding aspects of the dispreferred distribution, depending on the value of $\alpha$. Empirically, we observe that this strategy is not sensitive to the precise choice of $\alpha$ and a default value of $\alpha=0.5$ performs well across diverse text-to-image models.

Similar to~\cite{cfg}, the predicted noise of CHATS is combined as follows:
\begin{equation}
\label{eq:diff_sample}
\begin{split}
\tilde{\epsilon}_{\theta}(z_t, c) &= (1+s){\epsilon}_{\theta^+}(z_t, c) \\ 
&-s \Big[-\alpha {\epsilon}_{\theta^-}(z_t, c) + (1+\alpha) {\epsilon}_{\theta^-}(z_t)\Big].
\end{split}
\end{equation}

For flow matching models, the final expression remains similar, with $\epsilon_\theta$ simply replaced by $v_\theta$. To improve the sampling efficiency of our method, we introduce an approximation that constructs a \textit{proxy prompt} by linearly combining the prompt $c$ with a null prompt $\varnothing$ after they are transformed into text embeddings, i.e., $\hat c = -\alpha c + (1 + \alpha)\varnothing$. We then use this proxy prompt as the input to the dispreferred model:
\begin{equation}
\tilde{\epsilon}_{\theta}(z_t, c) = (1 + s)\epsilon_{\theta^+}(z_t, c)
- s \epsilon_{\theta^-}(z_t, \hat c).
\end{equation}

Note that when $\alpha = 0$, the method degenerates into CFG but utilizes two models. This modification significantly reduces the number of forward passes required for generation by unifying $c$ and $\varnothing$ into a single representation. Empirical evaluations demonstrate that this substitution achieves robust performance and serves as an effective surrogate for incorporating dispreferred information.

\begin{table*}[t]
 \centering
 \small
 \caption{HPS v2 results ($\uparrow$) on multiple benchmarks. The best results are in bold face.}
 \begin{tabular}{ll|cccccc}\toprule
 \textbf{Model} & \textbf{Method} & Anime & Concept-Art & Paintings & Photo & GenEval & DPG-Bench\\
 \midrule
 \multirow{3}{*}{SD1.5} 
    & Standard          & 19.66 & 18.11  & 18.25 & 20.13 & 21.20 & 18.88 \\
    & Diffusion-DPO        & 21.65 & 20.12  & 19.87 & 20.74 & 22.04 & 19.87 \\
    & CHATS & \textbf{27.74} & \textbf{26.17} & \textbf{26.04} & \textbf{26.43} & \textbf{26.17} & \textbf{24.97} \\
 \midrule
 \multirow{3}{*}{SDXL} 
    & Standard          & 30.20 & 28.38  & 28.19 & 26.88 & 28.19 & 27.38 \\
    & Diffusion-DPO        & 31.69 & 29.64  & 29.83 & 28.34 & 29.51 & 28.71 \\
    & CHATS & \textbf{32.96} & \textbf{30.94} & \textbf{31.08} & \textbf{29.62} & \textbf{29.81} & \textbf{29.51} \\
 \midrule
 \multirow{3}{*}{In-house T2I} 
    & Standard          & 25.00 & 23.60  & 23.18 & 26.09 & 27.23 & 25.68 \\
    & Diffusion-DPO     & 26.03 & 24.87  & 24.71 & 26.56 & 27.62 & 26.16 \\
    & CHATS       & \textbf{30.41} & \textbf{29.87} & \textbf{29.95} & \textbf{29.30} & \textbf{29.84} & \textbf{29.53} \\
 \bottomrule
 \end{tabular}
 \label{tab:main_exp}
\end{table*}

\section{Experiments}

\begin{table*}[t]
\centering
\small
 \setlength{\tabcolsep}{3pt}
\caption{Quantitative results (\%) averaged by 4 random seeds on GenEval.}
\label{tab:GENEVAL}
\begin{tabular}{llcccccc|c}\toprule
\textbf{Model} & \textbf{Method} 
             & Single object $\uparrow$
             & Two object $\uparrow$
             & Counting $\uparrow$
             & Colors $\uparrow$
             & Position $\uparrow$
             & Color attribution $\uparrow$
             & Overall $\uparrow$\\
\midrule
\multirow{3}{*}{SD1.5} 
    & Standard          & 95.31 & 38.13 & \phantom{0}5.94  & 75.27 & \phantom{0}7.25 & \phantom{0}9.25  & 38.52 \\
    & Diffusion-DPO     & 96.56 & \textbf{49.24} & \phantom{0}5.94  & \textbf{82.71} & \phantom{0}7.75 & \phantom{0}9.75  & 41.99 \\
    & CHATS             & \textbf{98.75} & 47.73 & \textbf{20.94} & 78.72 & \phantom{0}\textbf{7.75} & \textbf{10.75} & \textbf{44.11} \\
\midrule
\multirow{3}{*}{SDXL} 
    & Standard          & 98.44 & 68.43 & 40.62 & 83.78 & 10.25 & 20.75 & 53.71 \\
    & Diffusion-DPO     & \textbf{99.53} & \textbf{81.94} & 45.47 & \textbf{88.70} & 13.25 & 26.75 & 59.27 \\
    & CHATS             & 99.38 & 79.09 & \textbf{53.50} & 87.45 & \textbf{13.90} & \textbf{27.40} & \textbf{60.12} \\
\midrule
\multirow{3}{*}{In-house T2I} 
    & Standard          & 98.75 & 75.25 & 37.81 & 85.37 & 25.75 & 44.00 & 61.16 \\
    & Diffusion-DPO     & 99.06 & 75.25 & 40.00 & 83.78 & \textbf{27.25} & 48.00 & 62.22 \\
    & CHATS       & \textbf{99.06} & \textbf{82.58} & \textbf{47.50} & \textbf{90.16} & 26.50 & \textbf{52.75} & \textbf{66.42} \\
\bottomrule
\end{tabular}
\end{table*}

\begin{table*}[t]
\centering
\small
\caption{Quantitative results (\%) on DPG-Bench.}
\label{tab:dpgbench}
\begin{tabular}{llccccc|c}\toprule
\textbf{Model} & \textbf{Method} 
             & Global $\uparrow$
             & Entity $\uparrow$
             & Attribute $\uparrow$
             & Relation $\uparrow$
             & Other $\uparrow$
             & Overall $\uparrow$ \\
\midrule
\multirow{3}{*}{SD1.5} 
    & Standard          & 74.77 & 68.74  & 69.72  & 79.11  & 39.20  & 57.92  \\
    & Diffusion-DPO     & \textbf{81.76}  & 71.02  & 72.78  & \textbf{80.66}  & 38.40  & 61.67  \\
    & CHATS       & 77.51  & \textbf{73.67}  & \textbf{74.34}  & 80.23  & \textbf{58.80}  & \textbf{64.44}  \\
\midrule
\multirow{3}{*}{SDXL} 
    & Standard          & \textbf{85.41}  & 81.24  & 78.90  & 87.04  &  60.80 & 74.37  \\
    & Diffusion-DPO     & 84.50  & 82.00  & \textbf{80.16}  & 87.39  & 67.20  & 74.66  \\
    & CHATS       & 84.50  & \textbf{82.91}  & 80.12  & \textbf{87.47}  & \textbf{70.80}  & \textbf{76.08}  \\
\midrule
\multirow{3}{*}{In-house T2I} 
    & Standard          & 85.11  & 87.98  & 87.91 & 92.03  & 75.60  & 81.27  \\
    & Diffusion-DPO     & \textbf{86.63}  & 88.98  & 87.87  & 91.91  & 75.20 & 82.18  \\
    & CHATS       & 82.67  & \textbf{90.11}  & \textbf{88.17}  & \textbf{92.34}  & \textbf{84.40}  & \textbf{83.72}  \\
\bottomrule
\end{tabular}
\end{table*}

In this section, we first provide a detailed description of our experimental settings. We then compare our method with state-of-the-art approaches on multiple text-to-image generation models and benchmarks to highlight its effectiveness. Finally, we conduct essential ablation studies to investigate the impact of each component in our CHATS.

\subsection{Experimental Settings}

\textbf{Models and datasets.} We utilize two model families: diffusion models and flow matching models. For diffusion models, we employ Stable Diffusion 1.5 (SD1.5)~\cite{ldm} and Stable Diffusion XL-1.0 (SDXL)~\cite{sdxl}. For flow matching models, we deploy In-house T2I, a text-to-image generation model optimized for photorealistic images in e-commerce scenarios, based on the DiT architecture~\cite{dit}. We conduct experiments primarily on two preference optimization datasets, \textit{Pick-a-Pic v2 (PaP v2)}~\cite{pap} and \textit{OpenImagePreferences (OIP)}~\cite{oip} (see Supp.~\ref{sec:supp_exps_models_and_datasets} for more details). 

\textbf{Evaluation.} We utilize publicly available benchmark prompts from GenEval~\cite{geneval}, DPG-Bench~\cite{dpg_bench}, and HPS v2~\cite{hpsv2}. Detailed information about these benchmarks is provided in Supp.~\ref{sec:supp_exps_prompts}. We employ multiple evaluation metrics, including HPS v2~\cite{hpsv2}, ImageReward~\cite{image_reward}, and PickScore~\cite{pickscore}, which are trained on extensive preference datasets, offering robust insights about authentic human preferences. During sampling, by default we keep $s$ and $\alpha$ as 5 and 0.5, respectively.

\subsection{Main Results}

We compare our CHATS against two baselines: (1) the \textit{Standard} versions of SD1.5, SDXL, and In-house T2I, and (2) \textit{Diffusion-DPO}~\cite{diffusion-dpo}, the current state-of-the-art DPO method for text-to-image generation. We find that Diffusion-DPO performs better when finetuned on the PaP v2 dataset, whereas CHATS achieves superior results when trained on OIP. Consequently, by default we report results from Diffusion-DPO finetuned on PaP v2 alongside CHATS finetuned on OIP for a fair comparison. More details about data efficiency can be found in Sec.~\ref{sec:ablation} and Table~\ref{tab:dataset_analysis}.

In table~\ref{tab:main_exp} we report the HPS v2 results over diverse prompt groups as this aesthetic metric is better aligned with human judgment compared to ImageReward and PickScore metrics~\cite{hpsv2} (see Supp.~\ref{supp:full_aesthetic_results} for full results). CHATS consistently surpasses two baselines in all cases, underlining its robustness in handling various visual styles. Moreover, CHATS effectively narrows the aesthetic gaps among different models. For example, the largest discrepancy in average scores across the six groups occurs between SD1.5 and SDXL. Under the Standard setting, this gap is 8.83, but it decreases to 4.41 when adopting CHATS.

Table~\ref{tab:GENEVAL} presents the evaluation results on the GenEval benchmark, averaged over four random seeds. In this challenging scenario, CHATS outperforms baselines in 12 out of 18 tasks, achieving an average improvement of 2.39\% compared to Diffusion-DPO. Notably, CHATS excels in difficult tasks such as ``Counting'' (+10.18\% over Diffusion-DPO) and ``Color attribution'' (+2.13\%), underscoring its capability to capture complex relationships in multi-object or attribute-focused prompts. Furthermore, the results on DPG-Bench (Table~\ref{tab:dpgbench}) show a similar trend, with CHATS achieving the highest overall evaluation scores. Combined with the previous findings, these outcomes confirm the superiority of CHATS across diverse evaluation metrics and model architectures, effectively establishing a new state-of-the-art.

\subsection{Identifying Key Components of CHATS}
\label{sec:ablation}

\begin{table}
    \centering
    \small
    \setlength{\tabcolsep}{2pt}
    \caption{Ablation study on various configurations using In-house T2I with prompts from Photo. ``$s$'' denotes the guidance scalar in CFG as in Eq.~\ref{eq:cfg_score}.}
    \begin{tabular}{lc|c}
        \toprule
        \multicolumn{2}{c|}{Configuration} & \multirow{2}{*}{HPS v2 ($\uparrow$)}\\
        Finetuning & Sampling & \\
        \midrule
        single model (full data)      & $s$=5              & 28.64 \\
        single model (preferred data) & $s$=5              & 28.70 \\
        two models (w/o ref)          & $s$=5              & 28.99 \\
        two models                    & $s$=5              & 29.15 \\
        two models                    & $s$=5,$\alpha$=0.5 & \textbf{29.30} \\
        \bottomrule
    \end{tabular}
    \label{tab:modular_ablation}
\end{table}

\textbf{Modular ablation.} To elucidate the critical elements of our CHATS, we conduct an ablation study  by examining different fintuning and sampling configurations. The CHATS distinguishes itself from traditional finetuning methods through two primary innovations: (1) the introduction of two separate models dedicated to learning the preferred and dispreferred distributions, and (2) the integration of guidance signals from two models. 

As shown in Table~\ref{tab:modular_ablation}, the baseline configuration, referred to as ``single model (full data)'', trains a single model on the entire DPO dataset by converting all preference pairs ($z^+$, $z^-$, $c$) into individual samples ($z^+$, $c$) and ($z^-$, $c$). These samples are combined to construct a dataset without explicitly ranked preference pairs and are subsequently used to train the model with a standard flow matching loss~\cite{rectified_flow}. This configuration achieves an HPS v2 score of 28.64. Restricting the training to only preferred data results in a slight performance improvement to 28.70, suggesting that a single model fails to effectively leverage all preference information when full data is provided. Introducing a second model without a reference policy leads to a more substantial increase in performance, yielding an HPS v2 score of 28.99, thereby underscoring the advantages of independently modeling preferred and dispreferred distributions. By further integrating a reference model to guide the finetuning process elevates the score to 29.15. The most significant improvement is observed when incorporating the sampling strategy in CHATS with $\alpha=0.5$, resulting in the highest HPS v2 score of 29.30.

These findings demonstrate that both the dual-model architecture and the cooperative sampling strategy are crucial for enhancing the alignment of generated outputs with human preferences, thereby validating the critical design choices underpinning our CHATS.

\begin{table}[t]
    \centering
    \small
    \setlength{\tabcolsep}{4pt}
    \caption{Ablation studies on different finetuning datasets. HPS v2 results are reported for quantitative comparison.}
    \begin{tabular}{l|c|c|cc}
        \toprule
        \multirow{2}{*}{Method} & \multirow{2}{*}{Model} & \multirow{2}{*}{\begin{tabular}[c]{@{}l@{}}Finetuning\\ \; Dataset\end{tabular}} & \multicolumn{2}{c}{Benchmarks} \\
        & & & Photo & Paintings \\
        \midrule
        \multirow{4}{*}{Diffusion-DPO} & \multirow{2}{*}{SDXL} & PaP v2 & \textbf{28.34} & \textbf{29.83}  \\
         & & OIP & 28.16 & 29.03 \\
         \cline{2-5}
         & \multirow{2}{*}{In-house T2I} & PaP v2 & \textbf{26.56} & \textbf{24.71} \\
         & & OIP & 26.47 & 24.10 \\
         \midrule
        \multirow{4}{*}{CHATS} & \multirow{2}{*}{SDXL} & PaP v2 & 28.67 & 30.12 \\
         & & OIP & \textbf{29.62} & \textbf{31.08} \\
         \cline{2-5}
         & \multirow{2}{*}{In-house T2I} & PaP v2 & 28.73 & 28.70 \\
         & & OIP & \textbf{29.30} & \textbf{29.95}  \\
        \bottomrule
    \end{tabular}
    \label{tab:dataset_analysis}
\end{table}

\begin{table}[htbp!]
    \centering
    \small
    \caption{Throughput with 50 sampling steps, measured on NVIDIA A100 GPU with BF16 inference. ``bs'' specifies the batch size.}
    \begin{tabular}{l|c|c|c}
        \toprule
        \multirow{2}{*}{Model} & \multirow{2}{*}{Method} & \multicolumn{2}{c}{Throughput (img/sec)} \\
        \cline{3-4}
        & & bs=1 \phantom{00000} &bs=4 \phantom{00000} \\
        \midrule
        \multirow{2}{*}{SDXL} 
            & Standard      & \phantom{0}0.187 \phantom{(-11\%)} & 0.217 \phantom{(-3\%)}  \\
            & CHATS   & \phantom{0}0.166 (-11\%) & 0.210 (-3\%)  \\
        \midrule
        \multirow{2}{*}{In-house T2I} 
            & Standard      & 0.153 \phantom{(-3\%)} & 0.162 \phantom{(-2\%)} \\
            & CHATS   & 0.148 (-3\%) & 0.158 (-2\%) \\
        \bottomrule
    \end{tabular}
    \label{tab:throughput}
\end{table}

\begin{figure}[t]
 \centering
    \includegraphics[width=\columnwidth]{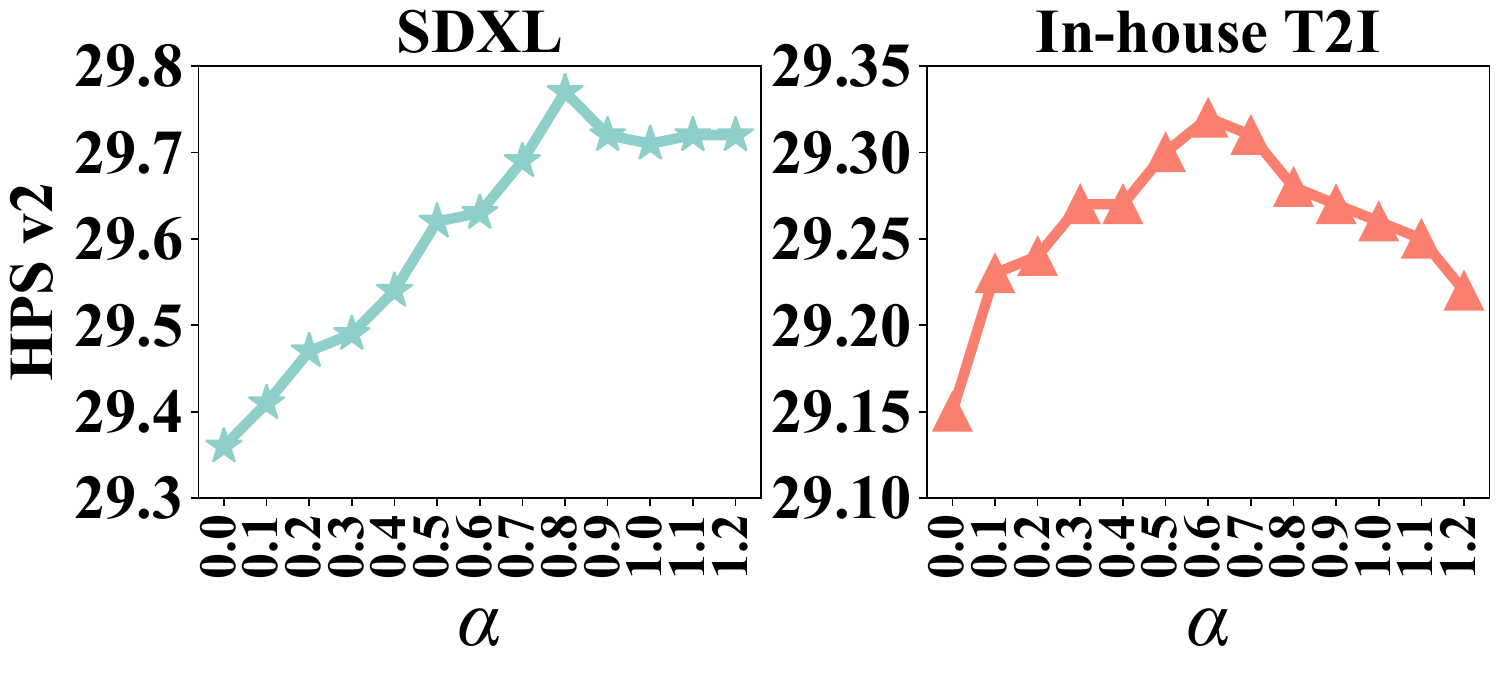}
    \caption{HPS v2 results by varying $\alpha$ of CHATS. The evaluation are conducted with SDXL and In-house T2I, using prompts from Photo.}
    \label{fig:alpha}
\end{figure}

\textbf{A small high-quality preference dataset is enough.} We further conduct ablation studies on two finetuning datasets with distinct characteristics: PaP v2~\cite{pap} and OIP~\cite{oip}. PaP v2 contains a large number of preference pairs but suffers from lower data quality, whereas OIP comprises $\thicksim1\%$ of PaP v2's entries yet features significantly higher-quality generated images.

As presented in Table~\ref{tab:dataset_analysis}, CHATS consistently outperforms Diffusion-DPO by a substantial margin across all scenarios. Notably, while Diffusion-DPO achieves superior performance when finetuned on PaP v2 dataset, CHATS excels when leveraging the smaller, high-quality OIP dataset. 

The contrasting behaviors between Diffusion-DPO and our CHATS stem from their distinct training objectives. Diffusion-DPO maximizes the difference in diffusion squared-loss values (cf. Eq.~\ref{eq:ddpm_obj}) from ranked preference pairs. However, in the OIP dataset, these differences are minimal because the ranked pairs are derived from two closely matched top-performing models. As a result, Diffusion-DPO struggles to distinguish and model preferred and dispreferred distributions, leading to suboptimal performance compared to training on the larger PaP v2 dataset with stronger preference signals.

In contrast, CHATS adopts a more direct approach by independently modeling the preferred and dispreferred distributions using two distinct models, while explicitly facilitating collaboration between them. This design enables CHATS to effectively utilize high-quality data and capture nuanced preference information, offering the possibility of performing preference optimization with only a small set of images.

\begin{figure}[t]
    \centering
    \includegraphics[width=.98\linewidth]{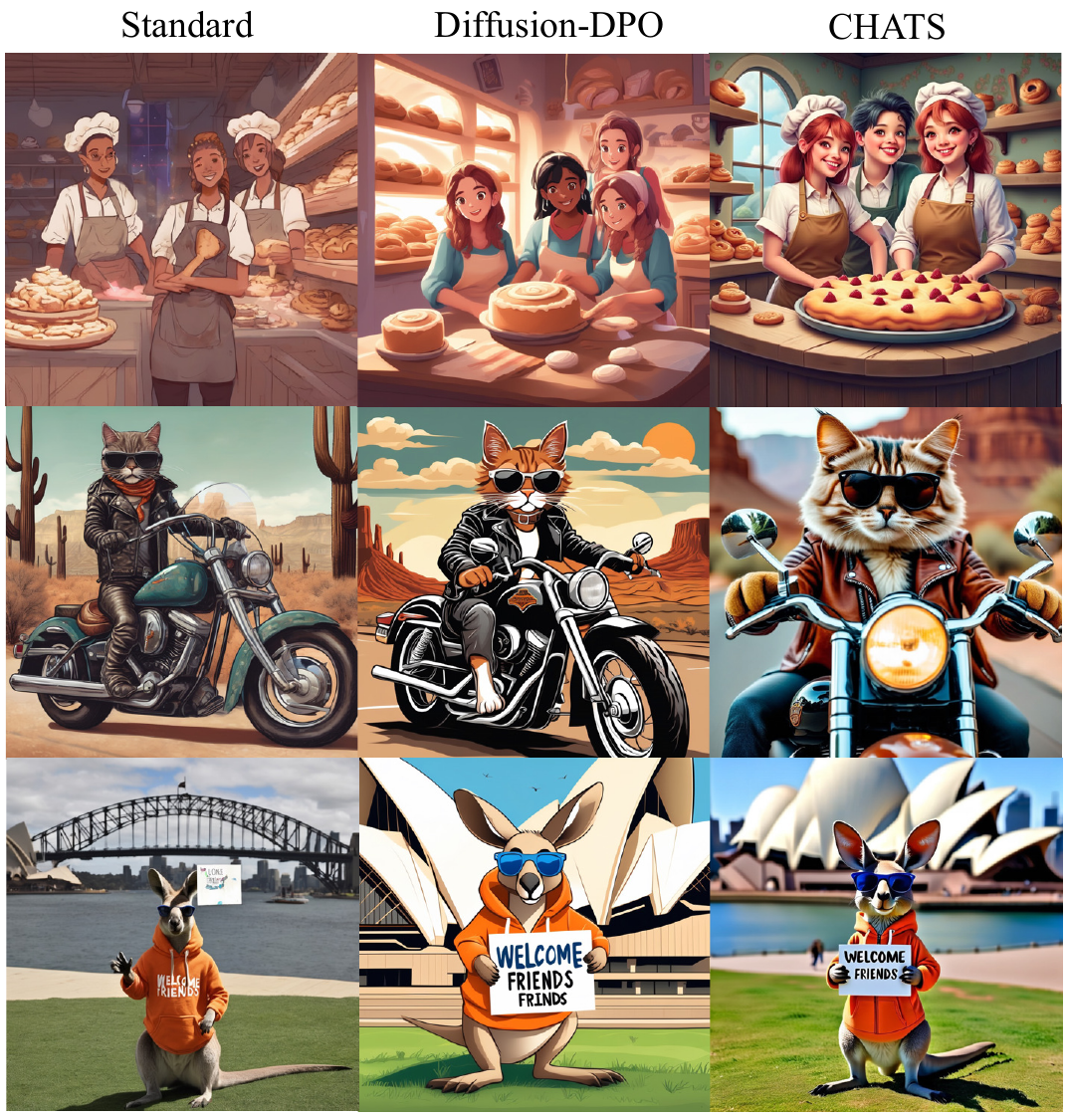}
    \caption{Qualitative comparison among various methods. The images generated by CHATS exhibit stronger text-image alignment, higher visual fidelity, and align more closely with human aesthetic standards. \textit{Prompts: 1) \underline{Three} friends working in a magical bakery. 2) An \underline{anthropomorphic} cat riding a Harley Davidson in Arizona with sunglasses and a leather jacket. 3) A kangaroo in an orange hoodie and blue sunglasses stands on the grass in front of the Sydney Opera House holding a sign that says \underline{Welcome Friends}.}}
    \label{fig:qualitative_results}
\end{figure}

\textbf{Robustness of $\alpha$.} The hyperparameter $\alpha$ in CHATS controls the balance between conditional and unconditional textual embeddings and regulates the contribution of the dispreferred distribution during sampling. We analyze its impact by varying $\alpha$. As shown in Fig.~\ref{fig:alpha} (see Supp.~\ref{supp:more_ablation_results} for more results), both SDXL and In-house T2I models exhibit consistent trends: as $\alpha$ increases from 0 to approximately 0.5, the HPS v2 results improve significantly. Beyond this range, the performance gradually saturates. Notably, compared to the naive CFG case ($\alpha = 0$), the proposed sampling strategy leveraging two models consistently enhances aesthetic scores across different model families. These results indicate that $\alpha$ is robust to variations in its value, with the default setting of $0.5$ performing reliably well.

\section{Conclusions and Limitations}

In this paper, we for the first time explored integrating human preference optimization and sampling process to advance text-to-image synthesis. We proposed CHATS, a novel training and sampling framework that enables better utilization of preference distributions by independently modeling preferred and dispreferred data. Our method exhibits high data efficiency, requiring only a small, high-quality dataset for finetuning to achieve strong results. Extensive experiments demonstrated the superiority of CHATS across various benchmarks, setting a new state-of-the-art compared to traditional non-synergistic baselines.

Our method has a limitation: in CFG, conditional and unconditional prompts are typically concatenated for a single forward pass with a batch size of 2, but our model processes them separately in two forward passes with a batch size of 1 each. Consequently, our approach slightly reduces inference throughput as shown in Table~\ref{tab:throughput}. However, this impact diminishes as the batch size increases. In industrial applications, the primary focus is on generation quality. In practice this additional overhead becomes negligible, as model distillation~\cite{disill_diffusion} can effectively unify the two models into a single, optimized model, such as FLUX.1-dev.

\section*{Impact Statement}
This paper presents work whose goal is to advance the field of machine learning in text-to-image generation. The robust generation capability of our method holds the potential to benefit domains such as design, data augmentation, and education, fostering interdisciplinary collaboration and inspiring creative exploration.

\bibliography{arxiv}
\bibliographystyle{tech_report}

\newpage
\appendix
\onecolumn
\section{Mathematical Derivations}
\label{supp:mathematical_derivations}

In this section, we present the mathematical derivation of the training objective for CHATS. First, we identify the global optimum of the target generative distribution $p_\theta(z_0|c)$ by maximizing the KL-divergence-constrained reward score in the framework of reinforcement learning from human feedback (RLHF). Subsequently, we utilize this global optimum to derive the training loss for our proposed method, ensuring a clear and rigorous formulation.

\subsection{Global Optimum of Reward Maximization}

RLHF refines a generative model $p_\theta (z_0|c)$ by maximizing the scores defined by a reward model $r$, simultaneously minimizing the KL-divergence between $p_\theta (z_0|c)$ and a reference distribution $p_\text{ref} (z_0|c)$ as:
\begin{equation}
    \max_{p_\theta} \mathbb E_{c \sim \mathcal D}\big[\mathbb E_{z_0 \sim  p_\theta(z_0|c)} r(z_0, c) - 
    \beta \mathbb D_{\text{KL}}
    \big(p_\theta(z_0|c) || p_{\text{ref}}(z_0|c)\big)\big]
\end{equation}
where $\mathcal D = \{c\}$ is the conditional dataset and $\beta$ is a scale scalar. Following~\citet{diffusion-dpo}, we have:
\begin{align}
&\max_{p_\theta} \mathbb E_{c \sim \mathcal D}\big[\mathbb E_{z_0 \sim  p_\theta(z_0|c)} r(z_0, c) - 
\beta \mathbb D_{\text{KL}} \big(p_\theta(z_0|c) || p_{\text{ref}}(z_0|c)\big)\big] \\
&=\max_{p_\theta} \mathbb E_{c \sim \mathcal D, z_0 \sim  p_\theta(z_0|c)} \big[r(z_0, c) - 
\beta \log \frac{p_\theta(z_0|c)}{p_\text{ref}(z_0|c)}\big] \\
&=\max_{p_\theta} \mathbb E_{c \sim \mathcal D, z_0 \sim  p_\theta(z_0|c)} \big[\mathbb E_{z_{1:T} \sim p_\theta(z_{1:T}|z_0, c)}\big(R(z_{0:T}, c) - 
\beta \log \frac{p_\theta(z_0|c)}{p_\text{ref}(z_0|c)}\big)\big] \\
&=\max_{p_\theta} \mathbb E_{c \sim \mathcal D, z_{0:T} \sim  p_\theta(z_{0:T}|c)} \big[R(z_{0:T}, c) - 
\beta \log \frac{p_\theta(z_0|c)}{p_\text{ref}(z_0|c)}\big] \\
&\geq  \max_{p_\theta} \mathbb E_{c \sim \mathcal D, z_{0:T} \sim  p_\theta(z_{0:T}|c)} \big[R(z_{0:T}, c) - 
\beta \log \frac{p_\theta(z_{0:T}|c)}{p_\text{ref}(z_{0:T}|c)}\big],
\end{align}
which is equivalent to:
\begin{equation}
\begin{split}
&\min_{p_\theta} \mathbb E_{c \sim \mathcal D, z_{0:T} \sim  p_\theta(z_{0:T}|c)} \big[ \log \frac{p_\theta(z_{0:T}|c)}{p_\text{ref}(z_{0:T}|c)} - \log e^{\frac{R(z_{0:T}, c)}{\beta}}\big]   \\
&=\min_{p_\theta} \mathbb E_{c \sim \mathcal D, z_{0:T} \sim  p_\theta(z_{0:T}|c)} \big[ 
\log \frac{ p_\theta(z_{0:T}|c)}{p_\text{ref}(z_{0:T}|c)e^{\frac{R(z_{0:T}, c)}{\beta}}/Z(c)} - \log Z(c)\big],
\label{eq:rlhf_obj}
\end{split}
\end{equation}
where $Z(c)= \sum_{z_{0:T}}p_\text{ref}(z_{0:T}|c)e^{\frac{R(z_{0:T}, c)}{\beta}}$ is the regularization term. We define:
\begin{equation}
    p^*(z_{0:T}|c) = p_\text{ref}(z_{0:T}|c)e^{\frac{R(z_{0:T}, c)}{\beta}}/Z(c),
\end{equation}
which is a valid distribution since $p^*(z_{0:T}|c) > 0$ for any $z_{0:T}$ and $\sum_{z_{0:T}} p^*(z_{0:T}|c)=1$. Therefore Eq.~\ref{eq:rlhf_obj} corresponds to:
\begin{align}
&\min_{p_\theta} \mathbb E_{c \sim \mathcal D, z_{0:T} \sim  p_\theta(z_{0:T}|c)} \big[ 
\log \frac{ p_\theta(z_{0:T}|c)}{p^*(z_{0:T}|c)} - \log Z(c)\big] \\
&=\min_{p_\theta} \mathbb E_{c \sim \mathcal D}\big[ \big(\mathbb E_{z_{0:T} \sim  p_\theta(z_{0:T}|c)} 
\log \frac{ p_\theta(z_{0:T}|c)}{p^*(z_{0:T}|c)}\big)  - \log Z(c)\big] \\
&= \min_{p_\theta} \mathbb E_{c \sim \mathcal D}\big[ \mathbb D_\text{KL} \big(p_\theta(z_{0:T}|c) || p^*(z_{0:T}|c)\big)  - \log Z(c)\big].
\end{align}

This optimization objective shows a global minimum as:
\begin{equation}
    p_\theta^*(z_{0:T}|c) = p^*(z_{0:T}|c) = p_\text{ref}(z_{0:T}|c)e^{\frac{R(z_{0:T}, c)}{\beta}}/Z(c).
\label{eq:global_optimum_of_r}
\end{equation}

\subsection{Training Objective of CHATS}
\label{sec:supp_training_obj_of_CHATS}

According to Eq.~\ref{eq:global_optimum_of_r}, the reward function can be reformulated as:
\begin{equation}
    R(z_{0:T},c) = \beta \log \frac{p_\theta^*(z_{0:T}|c)}{p_\text{ref}(z_{0:T}|c)} + \beta\log Z(c).
\label{eq:rf_reformulated}
\end{equation}
Hence the $r$ is expressed as:
\begin{align}
r(z_0, c) &= \mathbb{E}_{z_{1:T} \sim p_{\theta}(z_{1:T} | z_0, c)} \left[ R(z_{0:T}, c) \right] \\
&= \mathbb{E}_{z_{1:T} \sim p_{\theta}(z_{1:T} | z_0, c)} \left[ \beta \log \frac{p_\theta^*(z_{0:T}|c)}{p_\text{ref}(z_{0:T}|c)} + \beta\log Z(c) \right].
\label{eq:rlhf_r}
\end{align}
Since CHATS employs two different models for preferred and dispreferred distributions, parameterized by $\theta^+$ and $\theta^-$, the loss for each ranked pair can be recast by substituting Eq.~\ref{eq:rlhf_r} into Eq.~\ref{eq:twin_naive_obj}, omitting $c$, and setting $\beta = 1$ for simplicity, as follows:
\begin{equation}
\mathcal{L} (z_0^+, z_0^-) \!=\!
-\log \!\sigma\! \biggl(\!
{\mathbb{E}}_{z_{1:T}^+ \sim p_{\theta^+}(z_{1:T}^+|z_0^+), z_{1:T}^- \sim p_{\theta^-}(z_{1:T}^-|z_0^-)}\!\!
\left[ \log\! \frac{p_{\theta^+}(z_{0:T}^+)}{p_{\text{ref}}(z_{0:T}^+)}\! +\! \log\! \frac{p_{\theta^-}(z_{0:T}^-)}{p_{\text{ref}}(z_{0:T}^-)} + K\!\right]\!\biggr) ,
\end{equation}
where $K$ is a variable that depends only on $c$, and can therefore be treated as a constant during optimization. Consequently, the objective can be simplified as:
\begin{equation}
\mathcal{L} (z_0^+, z_0^-) \!=\!
-\log \!\sigma\! \biggl(\!
{\mathbb{E}}_{z_{1:T}^+ \sim p_{\theta^+}(z_{1:T}^+|z_0^+), z_{1:T}^- \sim p_{\theta^-}(z_{1:T}^-|z_0^-)}\!\!
\left[ \log\! \frac{p_{\theta^+}(z_{0:T}^+)}{p_{\text{ref}}(z_{0:T}^+)}\! +\! \log\! \frac{p_{\theta^-}(z_{0:T}^-)}{p_{\text{ref}}(z_{0:T}^-)}\!\right]\!\biggr) .
\end{equation}
Using $q(z_{1:T}|z_0)$to approximate $p_{\theta} (z_{1:T} | z_0)$ for both $\theta^+$and $\theta^-$, we have:
\begin{align}
\mathcal{L} (z_0^+, z_0^-) &= -\log\sigma [\mathbb{E}_{z_{1:T}^+ \sim q (z_{1:T}^+|z_0^+),z_{1:T}^- \sim q (z_{1:T}^-|z_0^-)} (\log \frac{p_{\theta^+}(z_{0:T}^+)}{p_{\text{ref}}(z_{0:T}^+)} + \log \frac{p_{\theta^-}(z_{0:T}^-)}{p_{\text{ref}}(z_{0:T}^-)})] \\
&= -\log\sigma [\mathbb{E}_{z_{1:T}^+ \sim q (z_{1:T}^+|z_0^+),z_{1:T}^- \sim q (z_{1:T}^-|z_0^-)} \big(\sum_{t=1}^T(\log \frac{p_{\theta^+}(z_{t-1}^+|z_t^+)}{p_{\text{ref}}(z_{t-1}^+|z_t^+)} + \log \frac{p_{\theta^-}(z_{t-1}^-|z_t^-)}{p_{\text{ref}}(z_{t-1}^-|z_t^-)})\big)] \\
&= -\log\sigma [\mathbb{E}_{z_{1:T}^+ \sim q (z_{1:T}^+|z_0^+),z_{1:T}^- \sim q (z_{1:T}^-|z_0^-)} \big(T \mathbb{E}_t(\log \frac{p_{\theta^+}(z_{t-1}^+|z_t^+)}{p_{\text{ref}}(z_{t-1}^+|z_t^+)} + \log \frac{p_{\theta^-}(z_{t-1}^-|z_t^-)}{p_{\text{ref}}(z_{t-1}^-|z_t^-)})\big)] \\
&= -\log\sigma [T \mathbb{E}_{\substack{(z_{t-1}^+, z_{t}^+) \sim q (z_{t-1}^+, z_{t}^+|z_0^+)\\(z_{t-1}^-, z_{t}^-) \sim q (z_{t-1}^-, z_{t}^-|z_0^-)}} \mathbb{E}_t(\log \frac{p_{\theta^+}(z_{t-1}^+|z_t^+)}{p_{\text{ref}}(z_{t-1}^+|z_t^+)} + \log \frac{p_{\theta^-}(z_{t-1}^-|z_t^-)}{p_{\text{ref}}(z_{t-1}^-|z_t^-)})] \\
&= -\log\sigma [T \mathbb{E}_t\mathbb{E}_{\substack{z_{t}^+ \sim q (z_{t}^+|z_0^+) \\ z_{t}^- \sim q (z_{t}^-|z_0^-)}} \mathbb {E}_{\substack{z_{t-1}^+ \sim q(z_{t-1}^+ | z_t^+, z_0^+)\\ z_{t-1}^- \sim q(z_{t-1}^- | z_t^-, z_0^-)}} (\log \frac{p_{\theta^+}(z_{t-1}^+|z_t^+)}{p_{\text{ref}}(z_{t-1}^+|z_t^+)} + \log \frac{p_{\theta^-}(z_{t-1}^-|z_t^-)}{p_{\text{ref}}(z_{t-1}^-|z_t^-)})] \\
&\leq -\mathbb{E}_{t,\substack{z_{t}^+ \sim q (z_{t}^+|z_0^+) \\ z_{t}^- \sim q (z_{t}^-|z_0^-)}} \log\sigma [ T\mathbb {E}_{\substack{z_{t-1}^+ \sim q(z_{t-1}^+ | z_t^+, z_0^+) \\ z_{t-1}^- \sim q(z_{t-1}^- | z_t^-, z_0^-)}} (\log \frac{p_{\theta^+}(z_{t-1}^+|z_t^+)}{p_{\text{ref}}(z_{t-1}^+|z_t^+)} + \log \frac{p_{\theta^-}(z_{t-1}^-|z_t^-)}{p_{\text{ref}}(z_{t-1}^-|z_t^-)})] \\
&= -\mathbb{E}_{t, \substack{z_{t}^+ \sim q (z_{t}^+|z_0^+) \\ z_{t}^- \sim q (z_{t}^-|z_0^-)}} \log\sigma [ - T \bigg(\mathbb {E}_{\substack{z_{t-1}^+ \sim q(z_{t-1}^+ | z_t^+, z_0^+)\\ z_{t-1}^- \sim q(z_{t-1}^- | z_t^-, z_0^-)}} \big( \notag \\
& + \log \frac{q(z_{t-1}^+|z_t^+, z_0^+)}{p_{\theta^+}(z_{t-1}^+|z_t^+)} - \log \frac{q(z_{t-1}^+|z_t^+, z_0^+)}{p_\text{ref}(z_{t-1}^+|z_t^+)} + \log \frac{q(z_{t-1}^-|z_t^-, z_0^-)}{p_{\theta^-}(z_{t-1}^-|z_t^-)} - \log \frac{q(z_{t-1}^-|z_t^-, z_0^-)}{p_\text{ref}(z_{t-1}^-|z_t^-)} \big) \bigg)] \\
&= -\mathbb{E}_{t, \substack{z_{t}^+ \sim q (z_{t}^+|z_0^+) \\ z_{t}^- \sim q (z_{t}^-|z_0^-)}} \log\sigma [ - T \bigg( \notag \\
& + \mathbb{D}_{\text{KL}} \big( q(z_{t-1}^+|z_{t}^+,z_{0}^+) || p_{\theta^+}(z_{t-1}^+|z_{t}^+)  \big) - \mathbb{D}_{\text{KL}} \big( q(z_{t-1}^+|z_{t}^+,z_{0}^+) || p_{\text{ref}}(z_{t-1}^+|z_{t}^+)  \big) \notag \\
&+ \mathbb{D}_{\text{KL}} \big( q(z_{t-1}^-|z_{t}^-,z_{0}^-) || p_{\theta^-}(z_{t-1}^-|z_{t}^-) \big) - \mathbb{D}_{\text{KL}} \big( q(z_{t-1}^-|z_{t}^-,z_{0}^-) || p_{\text{ref}}(z_{t-1}^-|z_{t}^-)  \big)  \bigg)].
\label{eq:chats_derivation}
\end{align}

For diffusion models, minimizing $\mathbb{D}_{\text{KL}} \big( q(z_{t-1}|z_{t},z_{0}) \,\|\, p_{\theta}(z_{t-1}|z_{t}) \big)$ is equivalent to minimizing $\| \epsilon - \epsilon_\theta(z_t, t) \|^2$. Accordingly, the training loss of CHATS is formulated as:
\begin{equation}
\begin{split}
\mathcal{L}_\text{CHATS}^{\text{Diffusion}} (z_0^+, z_0^-) &= -\mathbb{E}_{t\sim \mathcal U(0,T), \substack{z_{t}^+ \sim q (z_{t}^+|z_0^+) \\ z_{t}^- \sim q (z_{t}^-|z_0^-)}}
\log\sigma \bigg[-T \bigg( \\
&+ \big\|\epsilon^+ \!\!-\! \epsilon_{\theta^+} (z^+_t, t)\big\|^2 \!\!-\! \big\|\epsilon^+ - \epsilon_{\text{ref}} (z^+_t, t)\big\|^2 + \big\|\epsilon^- \!\!-\! \epsilon_{\theta^-} (z^-_t, t)\big\|^2 \!\!-\! \big\|\epsilon^- - \epsilon_{\text{ref}} (z^-_t, t)\big\|^2 \!\bigg) \bigg]\!,
\end{split}
\end{equation}
where $\epsilon^+,\epsilon^+ \sim \mathcal{N}(\mathbf{0}, \mathbf{I})$ and $z_t^+, z_t^-$ are defined as in Eq.~\ref{eq:ddpm_to_xt}.

For flow matching models, the minimization of $\mathbb{D}_{\text{KL}} \big( q(z_{t-1}|z_{t},z_{0}) || p_{\theta}(z_{t-1}|z_{t}) \big)$ is equivalent to the minimization of $||v - v_\theta(z_t, t)||^2$. In this case, the loss function is defined as:
\begin{equation}
\begin{split}
\mathcal{L}_\text{CHATS}^{\text{Flow}} (z_0^+, z_0^-) &= -\mathbb{E}_{t\sim \mathcal U(0,T), \substack{z_{t}^+ \sim q (z_{t}^+|z_0^+) \\ z_{t}^- \sim q (z_{t}^-|z_0^-)}}
\log\sigma \bigg[-T \bigg( \\
&+ \big\|v^+ \!\!-\! v_{\theta^+} (z^+_t, t)\big\|^2 \!\!-\! \big\|v^+ - v_{\text{ref}} (z^+_t, t)\big\|^2 + \big\|v^- \!\!-\! v_{\theta^-} (z^-_t, t)\big\|^2 \!\!-\! \big\|v^- - v_{\text{ref}} (z^-_t, t)\big\|^2 \!\bigg) \bigg]\!,
\end{split}
\end{equation}
where $z_t^+$ and $z_t^-$ are defined as the linear interpolation between noise $\epsilon \sim \mathcal{N}(\mathbf{0}, \mathbf{I})$ and the data samples $z_0^+$ or $z_0^-$, following Eq.~\ref{eq:flow_xt}.

\subsection{Convergence Properties of CHATS}

Given that DPO is invariant to affine transformations of the reward, for reward: $R'(z_{0:T}, c) = a \cdot R(z_{0:T}, c) + b$, the optimal policy becomes (cf. Eq.~\ref{eq:global_optimum_of_r}):
\begin{equation}
p^*(z_{0:T}\mid c) = \frac{p_{\mathrm{ref}}(z_{0:T}\mid c) e^{a \cdot R(z_{0:T}, c) + b}}{Z'(c)}.
\end{equation}

CHATS decomposes reward of traditional DPO (Eq.~\ref{eq:rf_reformulated}) into two parts:
\begin{equation}
R^+(\theta^+) = \log\frac{p_{\theta^+}(z_{0:T}^+\mid c)}{p_{\mathrm{ref}}(z_{0:T}^+\mid c)}, \quad
R^-(\theta^-) = \log\frac{p_{\theta^-}(z_{0:T}^-\mid c)}{p_{\mathrm{ref}}(z_{0:T}^-\mid c)},
\end{equation}
with $\beta$ and $\log Z(c)$ omitted since they are constants for optimization. Defining the \textit{effective reward} as: $R_{\mathrm{CHATS}} = R^+(\theta^+) + R^-(\theta^-)$, the resulting joint optimal distribution becomes:
\begin{equation}
p^*(z_{0:T}\mid c) = \frac{p_{\mathrm{ref}}(z_{0:T}\mid c) e^{R_{\mathrm{CHATS}}}}{Z^{\text{CHATS}}(c)}.
\end{equation}

Under the assumption of $L$-smoothness and using standard gradient descent, we obtain the recursive inequality:
\begin{equation}
    \mathcal{L}_{k+1} \le \mathcal{L}_k - \frac{\eta}{2} \|\nabla \mathcal{L}(\theta^+_k, \theta^-_k)\|^2,
\end{equation}
which ensures the CHATS loss (Eq.~\ref{eq:twin_naive_obj}) decreases monotonically and converges. Since their combination recovers the same optimal joint distribution as traditional DPO methods~\cite{diffusion-dpo}, CHATS preserves theoretical foundations of DPO.

\subsection{Theoretical Foundations of CHATS Sampler}
\label{sec:chats_sampler}

Start from Bayes’ rule for a classifier:
\begin{equation}
    p(c \mid z_t) = \frac{p(z_t \mid c)p(c)}{p(z_t)},
\end{equation}
since $p(c)$ can be regarded as a constant during optimization, CFG defines the guided distribution by raising $p(c \mid z_t)$ with a guidance scale $s$:

\begin{equation}
\tilde{p}(z_t \mid c) \propto p(z_t|c)\bigl[p(c \mid z_t)\bigr]^s.
\end{equation}

Substitute the expression for $p(c \mid z_t)$ and omit $p(c)$ yields:

\begin{equation}
\tilde{p}(z_t \mid c) \propto p(z_t \mid c)^{1+s}p(z_t)^{-s}.
\end{equation}

In CHATS, two models are used:
\begin{itemize}
    \item The preferred model $p_{\theta^+}(z_t \mid c)$,
    \item The dispreferred model $p_{\theta^-}(z_t \mid c)$ (with its unconditional form $p_{\theta^-}(z_t)$). 
\end{itemize}

For each model, we can write a classifier-like term via Bayes’ rule. For the preferred model:

\begin{equation}
    p_{\theta^+}(c \mid z_t) = \frac{p_{\theta^+}(z_t \mid c)p(c)}{p_{\theta^+}(z_t)},
\end{equation}

and for the dispreferred model:

\begin{equation}
p_{\theta^-}(c \mid z_t) = \frac{p_{\theta^-}(z_t \mid c)p(c)}{p_{\theta^-}(z_t)}.
\end{equation}

Assuming $p_{\theta^+}(z_t) \approx p_{\theta^-}(z_t)$, we combine the two signals by defining a composite log-odds score:

\begin{equation}
\Delta(z_t,c)=\log\frac{p_{\theta^+}(z_t \mid c)}{p_{\theta^-}(z_t)}+\alpha\log\frac{p_{\theta^-}(z_t \mid c)}{p_{\theta^-}(z_t)}.
\end{equation}

The first term tends to generate features favored by the preferred model while suppressing the background features typically produced by the dispreferred model in its unconditional output (similar to CFG), and the second term further accounts for the shift in the output of the dispreferred model when conditioned on $c$, with its impact regulated by a scalar $\alpha$. In this form, the useful information in $p_{\theta^-} (z_t \mid c)$ is effectively utilized as well.

Following CFG, we define the CHATS guided distribution as:

\begin{equation}
\tilde{p}_\theta(z_t \mid c) \propto p_{\theta^+}(z_t \mid c)\exp\Bigl(s\cdot\Delta(z_t,c)\Bigr).
\end{equation}

Substitute $\Delta(z_t,c)$:

\begin{equation}
\tilde{p}_\theta(z_t \mid c) \propto p_{\theta^+}(z_t \mid c) \cdot \exp\left(s\left[\log\frac{p_{\theta^+}(z_t \mid c)}{p_{\theta^-}(z_t)}+\alpha\cdot\log\frac{p_{\theta^-}(z_t \mid c)}{p_{\theta^-}(z_t)}\right]\right).
\end{equation}

Using $\exp(s\log A)=A^s$, we have

\begin{equation}
\tilde{p}_\theta(z_t \mid c) \propto p_{\theta^+}(z_t \mid c)
\left(\frac{p_{\theta^+}(z_t \mid c)}{p_{\theta^-}(z_t)}\right)^s
\left(\frac{p_{\theta^-}(z_t \mid c)}{p_{\theta^-}(z_t)}\right)^{\alpha s}.
\end{equation}

Grouping terms, we obtain

\begin{equation}
\tilde{p}_\theta(z_t \mid c) \propto p_{\theta^+}(z_t \mid c)^{1+s}p_{\theta^-}(z_t \mid c)^{\alpha s}p_{\theta^-}(z_t)^{-(1+\alpha)s},
\end{equation}

which is the same with Eq.~\ref{eq:twin_sampling}. The final guided distribution is not merely a sharpened version of $p_{\theta^+}(z_t \mid c)$. It also leverages the dispreferred model. The term$\left(\frac{p_{\theta^-}(z_t \mid c)}{p_{\theta^-}(z_t)}\right)^{\alpha s}$ adjusts the output based on how conditioning on $c$ changes the dispreferred model’s behavior. This derivation, starting from $p(c\mid z_t)$ for both models, provides a theoretical foundation for the CHATS sampling distribution analogous to that of CFG.

\section{More Related Works}
\label{supp:more_related_works}

We note that a recent method Diffusion-NPO~\cite{diffusion_npo} also employs two models for preference optimization. However, CHATS diverges from this approach in several important respects. First, CHATS is motivated by the observation that no existing approach integrates preference optimization and test time sampling within a unified framework to exploit both positive and negative signals, whereas Diffusion-NPO seeks to address limitations of diffusion models by training a dedicated negative preference model to handle unconditional outputs. Second, Diffusion-NPO trains its models sequentially, first fitting a standard DPO-optimized model and then training a negative preference model on inverted preference data. In contrast, CHATS \textit{jointly optimizes both preferred and dispreferred distributions under a single objective}, eliminating sequential steps and yielding significantly higher training efficiency. Moreover, CHATS demonstrates superior data efficiency by requiring only approximately 7K preference pairs. Finally, at inference CHATS employs a three term guidance scheme grounded in rigorous theoretical analysis (see Sec.~\ref{sec:chats_sampler}) that combines outputs from both models to guide generation toward preferred modes and repel undesirable ones. Diffusion-NPO relies instead on a two term guidance scheme that substitutes the unconditional predictor with the negative preference model. By integrating preference signals consistently during both training and inference, CHATS achieves more coherent alignment and markedly improved image quality.

\section{Full Experimental Settings}
\label{supp:full_experimental_settings}

\subsection{Models and Finetuning Datasets}
\label{sec:supp_exps_models_and_datasets}

We utilize two model families: diffusion models and flow matching models. For diffusion models, we employ Stable Diffusion 1.5 (SD1.5)~\cite{ldm} and Stable Diffusion XL-1.0 (SDXL)~\cite{sdxl}. For flow matching models, we deploy In-house T2I, a text-to-image generation model optimized for photorealistic images in e-commerce scenarios, based on the DiT architecture~\cite{dit}.

We finetune our models on two preference optimization datasets. The first, \textit{Pick-a-Pic v2 (PaP v2)}~\cite{pap}, comprises 851,293 tie-free preference pairs collected through user interactions with SD2.1, SD1.5 variants, and SDXL variants~\cite{ldm} under various CFG values. The second, \textit{OpenImagePreferences (OIP)}~\cite{oip}, is a recently released high-quality image preference dataset containing 7,459 preference pairs of 1024$\times$1024 images generated by SD3.5-large~\cite{sd3} and FLUX.1-dev~\cite{fluxdev}. Compared to PaP v2, OIP offers superior image quality, including enhanced resolution and textual fidelity. We leverage both datasets to demonstrate the distinct advantages of our proposed method.

\subsection{Finetuning Details}

Following the approach from~\citet{diffusion-dpo}, we employ Adafactor~\cite{adafactor} to finetune the SDXL model and AdamW~\cite{adamw} for the SD1.5 and In-house T2I models. Training is conducted with an effective batch size of 512, maintaining an image resolution of 1024. The default learning rate is set to $1 \times 10^{-8}$, and a learning rate scaling strategy based on batch size increases is utilized to accelerate the finetuning. $T$ (cf. Eq.~\ref{eq:diffusion_chat_obj} and Eq.~\ref{eq:flow_chat_obj}) is fixed as 1000.

\subsection{Evaluation Prompts}
\label{sec:supp_exps_prompts}

\textbf{HPS v2.} The HPS v2~\cite{hpsv2} is a comprehensive benchmark for aesthetic evaluation, comprising 3,200 prompts evenly distributed across four distinct styles: ``Anime'',  ``Concept-Art'', ``Paintings'', and ``Photo'' with 800 prompts per style. We select this benchmark due to its extensive number of prompts, which ensures the stability of results and strengthens the reliability of the evaluation.

\textbf{GenEval.} GenEval~\cite{geneval} is an object-centric evaluation framework designed to assess text-to-image models on compositional image attributes, such as object co-occurrence, spatial positioning, quantity, and color. It comprises 553 prompts and is typically used in conjunction with object detection models to report compositional scores.

\textbf{DPG-Bench.} DPG-Bench~\cite{dpg_bench} is a challenging benchmark consisting of 1,065 lengthy and dense prompts, with each describing multiple objects characterized by a wide range of attributes and complex relationships. 

\subsection{Evaluation Metrics}

\begin{table*}[t]
 \centering
 \small
 \caption{Full aesthetic results ($\uparrow$) on ``Anime'' and ``Concept-Art''.}
 \begin{tabular}{ll|ccc|ccc}\toprule
 \textbf{Model} & \textbf{Method} & \multicolumn{3}{c}{Anime} & \multicolumn{3}{c}{Concept-Art}\\
 & & ImageReward & HPS v2 & PickScore & ImageReward & HPS v2 & PickScore \\
 \midrule
 \multirow{3}{*}{SD1.5} 
    & Standard          & -134.51 & 19.66 & 19.46 & -121.69 & 18.11 & 19.47 \\
    & Diffusion-DPO     & \phantom{0}-95.95 & 21.65 & 20.13 & \phantom{0}-89.71 & 20.12 & 20.14 \\
    & CHATS             & \textbf{\phantom{-0}34.35} & \textbf{27.74} & \textbf{21.38} & \textbf{\phantom{-0}30.74} & \textbf{26.17} & \textbf{20.80} \\
 \midrule
 \multirow{3}{*}{SDXL} 
    & Standard          & \phantom{0}94.99 & 30.20 & 22.80 & \phantom{0}90.24 & 28.38 & 22.19 \\
    & Diffusion-DPO     & 111.73 & 31.69 & 23.09 & \phantom{0}98.26  & 29.64 & 22.27 \\
    & CHATS             & \textbf{122.73} & \textbf{32.96} & \textbf{23.17} & \textbf{107.21} & \textbf{30.94} & \textbf{22.38} \\
 \midrule
 \multirow{3}{*}{In-house T2I} 
    & Standard          & \phantom{0}49.23  & 25.00 & 20.99 & \phantom{0}53.89  & 23.60 & 20.50 \\
    & Diffusion-DPO     & \phantom{0}65.49  & 26.03 & 21.27 & \phantom{0}69.82  & 24.87 & 20.83 \\
    & CHATS             & \textbf{112.41} & \textbf{30.41} & \textbf{22.13} & \textbf{114.07} & \textbf{29.87} & \textbf{21.66}  \\
 \bottomrule
 \end{tabular}
 \label{tab:supp_anime_and_concept}
\end{table*}

\begin{table*}[t]
 \centering
 \small
 \caption{Full aesthetic results ($\uparrow$) on ``Paintings'' and ``Photo''.}
 \begin{tabular}{ll|ccc|ccc}\toprule
 \textbf{Model} & \textbf{Method} & \multicolumn{3}{c}{Paintings} & \multicolumn{3}{c}{Photo}\\
 & & ImageReward & HPS v2 & PickScore & ImageReward & HPS v2 & PickScore \\
 \midrule
 \multirow{3}{*}{SD1.5} 
    & Standard          &  -123.31 & 18.25 & 19.53 & -84.69 & 20.13 & 20.19\\ 
    & Diffusion-DPO     &  \phantom{0}-97.24  & 19.87 & 20.08 & -72.05 & 20.74 & 20.47\\
    & CHATS             &  \textbf{\phantom{-0}40.53}   & \textbf{26.04} & \textbf{20.94} & \textbf{\phantom{-}18.71}  & \textbf{26.43} & \textbf{21.44} \\
 \midrule
 \multirow{3}{*}{SDXL} 
    & Standard          & \phantom{0}91.39 & 28.19 & 22.31 & 67.31 & 26.88 & 22.26 \\
    & Diffusion-DPO     & 108.19 & 29.83 & 22.42 & 82.89 & 28.34 & 22.51 \\
    & CHATS             & \textbf{114.10} & \textbf{31.08} & \textbf{22.55} & \textbf{88.72} & \textbf{29.62} & \textbf{22.61}  \\
 \midrule
 \multirow{3}{*}{In-house T2I} 
    & Standard          & \phantom{0}53.03  & 23.18 & 20.48 & 73.00 & 26.09 & 21.86 \\
    & Diffusion-DPO     & \phantom{0}70.20  & 24.71 & 20.88 & 80.51 & 26.56 & 21.98 \\
    & CHATS             & \textbf{116.36} & \textbf{29.95} & \textbf{21.54} & \textbf{97.02} & \textbf{29.30} & \textbf{22.17} \\
 \bottomrule
 \end{tabular}
 \label{tab:supp_paintings_and_photo}
\end{table*}

\begin{table*}[t]
 \centering
 \small
 \caption{Full aesthetic results ($\uparrow$) on ``GenEval'' and ``DPG-Bench''.}
 \begin{tabular}{ll|ccc|ccc}\toprule
 \textbf{Model} & \textbf{Method} & \multicolumn{3}{c}{GenEval} & \multicolumn{3}{c}{DPG-Bench}\\
 & & ImageReward & HPS v2 & PickScore & ImageReward & HPS v2 & PickScore \\
 \midrule
 \multirow{3}{*}{SD1.5} 
    & Standard          & -111.58 & 21.20 & 20.46 & -117.39 & 18.88 & 19.70 \\ 
    & Diffusion-DPO     & \phantom{0}-95.61  & 22.04 & 20.72 & -100.05 & 19.87 & 20.01 \\
    & CHATS             & \textbf{\phantom{0}-19.74}  & \textbf{26.17} & \textbf{21.55} & \textbf{\phantom{0}-21.47} & \textbf{24.97} & \textbf{20.79} \\
 \midrule
 \multirow{3}{*}{SDXL} 
    & Standard          & \phantom{-0}55.14   & 28.19 & 22.65 & \phantom{-0}48.84 & 27.38 & 21.99 \\
    & Diffusion-DPO     & \phantom{-0}79.19   & 29.51 & 22.94 & \phantom{-0}57.17 & 28.71 & 22.06 \\
    & CHATS             & \textbf{\phantom{-0}84.72}   & \textbf{29.81} & \textbf{22.96} & \textbf{\phantom{-0}65.79} & \textbf{29.51} & \textbf{22.08} \\
 \midrule
 \multirow{3}{*}{In-house T2I} 
    & Standard          & \phantom{-0}76.61  & 27.23 & 22.13 & \phantom{-0}56.44 & 25.68 & 21.08 \\
    & Diffusion-DPO     & \phantom{-0}84.61	 & 27.62 & 22.24 & \phantom{-0}60.73 & 26.16 & 21.21\\
    & CHATS             & \textbf{\phantom{-}105.31} & \textbf{29.84} & \textbf{22.54} & \textbf{\phantom{-0}89.28} & \textbf{29.53} & \textbf{21.66} \\
 \bottomrule
 \end{tabular}
 \label{tab:supp_geneval_and_dpg}
\end{table*}

For aesthetic evaluation, we employ three aesthetic evaluators:

\textbf{HPS v2.} HPS v2~\cite{hpsv2} is built upon a CLIP (ViT-H/14) model, finetuned on the Human Preference Datasets v2. This dataset includes 798,090 human preference annotations across 433,760 image pairs, enabling HPS v2 to accurately predict human preferences for images generated from textual prompts.

\textbf{ImageReward.} ImageReward~\cite{image_reward} is the first general-purpose human preference scoring model for text-to-image generation. It is trained using a systematic annotation pipeline, incorporating both rating and ranking methodologies, and has collected over 137,000 expert comparisons.

\textbf{PickScore.} PickScore~\cite{pickscore} is another preference evaluator, finetuned on the Pick-a-Pic dataset, a large open dataset comprising text-to-image prompts and real user preferences for generated images.

Among these three evaluators, HPS v2 demonstrates the highest correspondence with human annotators. Therefore, we use it as the primary evaluator for aesthetic evaluation in the main text. In the following section, we provide the complete results from all three evaluators.

\section{Full Aesthetic Results}
\label{supp:full_aesthetic_results}

We report the full aesthetic results on Table~\ref{tab:supp_anime_and_concept},~\ref{tab:supp_paintings_and_photo} and~\ref{tab:supp_geneval_and_dpg}. Consistent with the findings presented in the main text, our CHATS outperforms two baseline methods across all aesthetic evaluators and all groups of prompts. This further highlights the effectiveness of CHATS.

\section{More Ablation on $\alpha$}
\label{supp:more_ablation_results}

\begin{figure}[t]
    \centering
    \subfigure[Anime]{\includegraphics[width=0.48\textwidth]{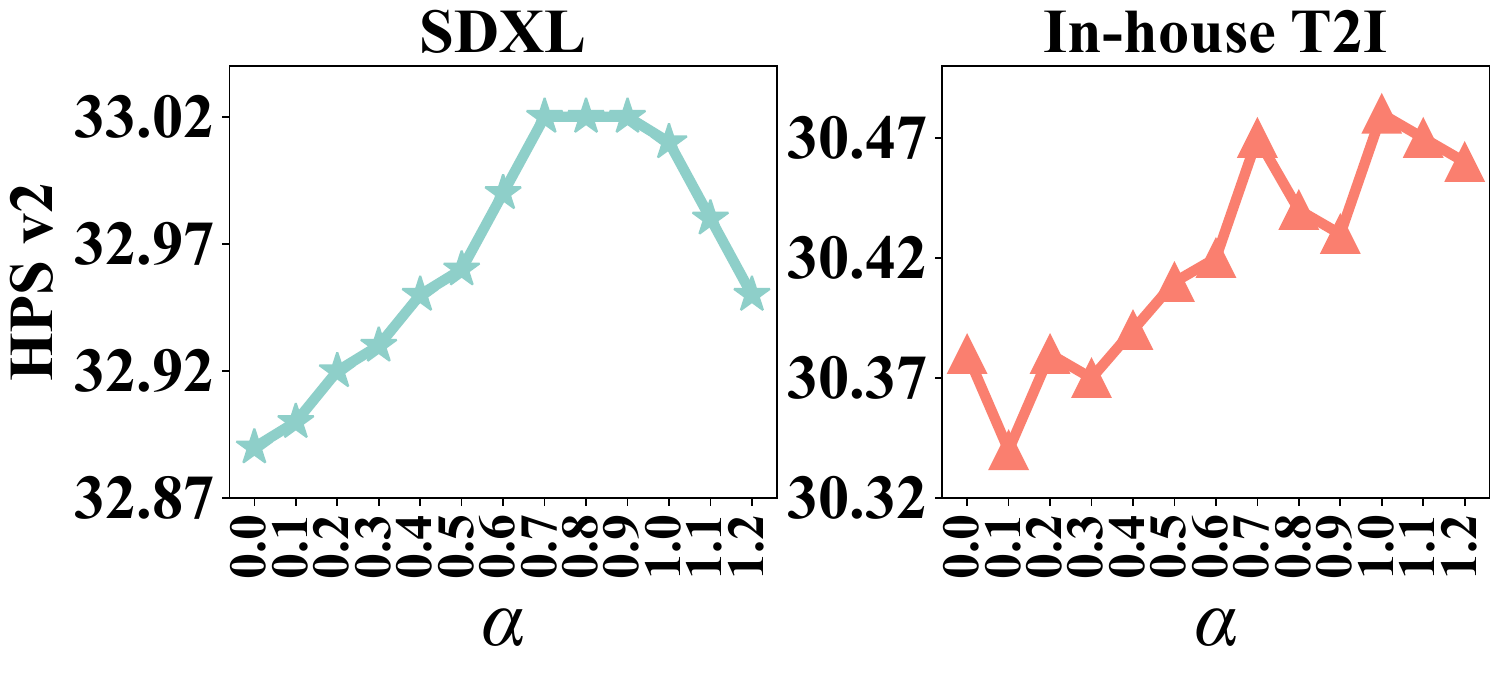} \label{fig:supp_alpha_anime}}
    \hfill
    \subfigure[Concept-Art]{\includegraphics[width=0.48\textwidth]{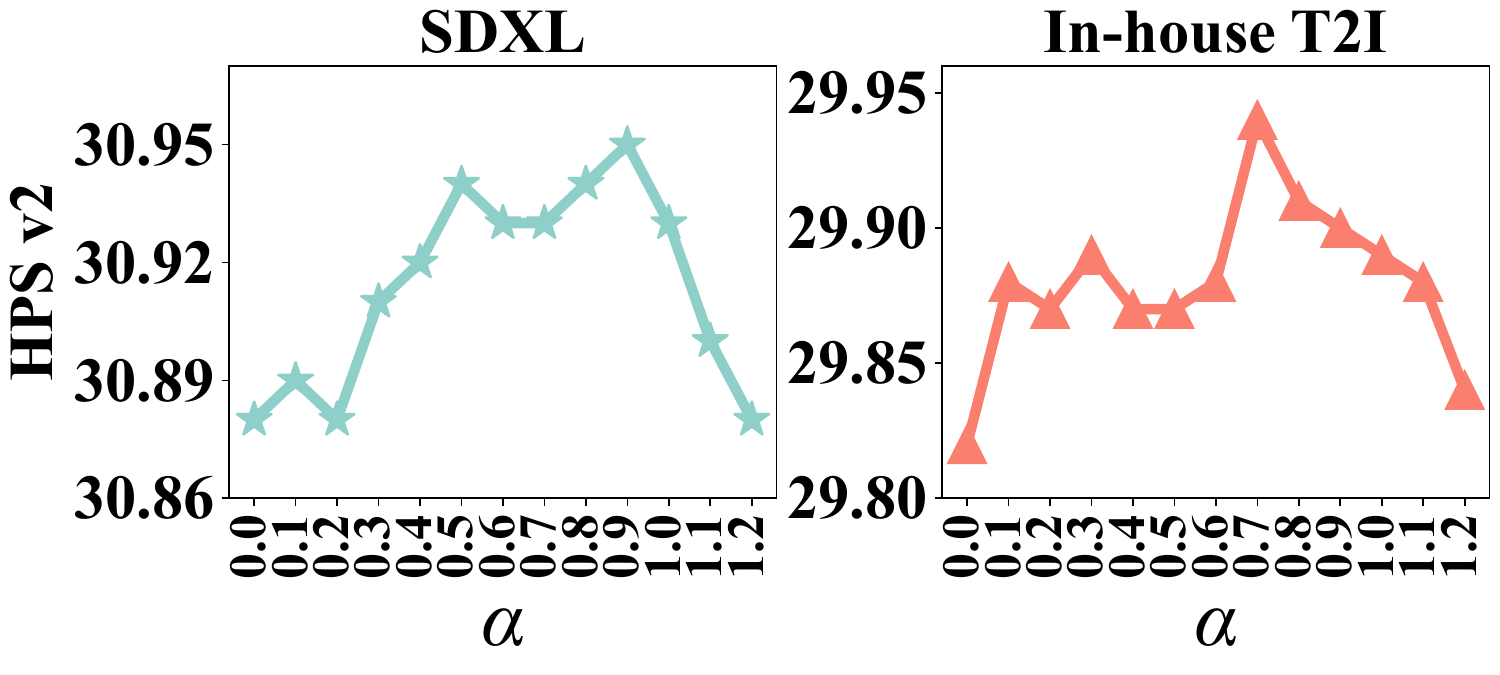} \label{fig:supp_alpha_concept}}
    \subfigure[Paintings]{\includegraphics[width=0.48\textwidth]{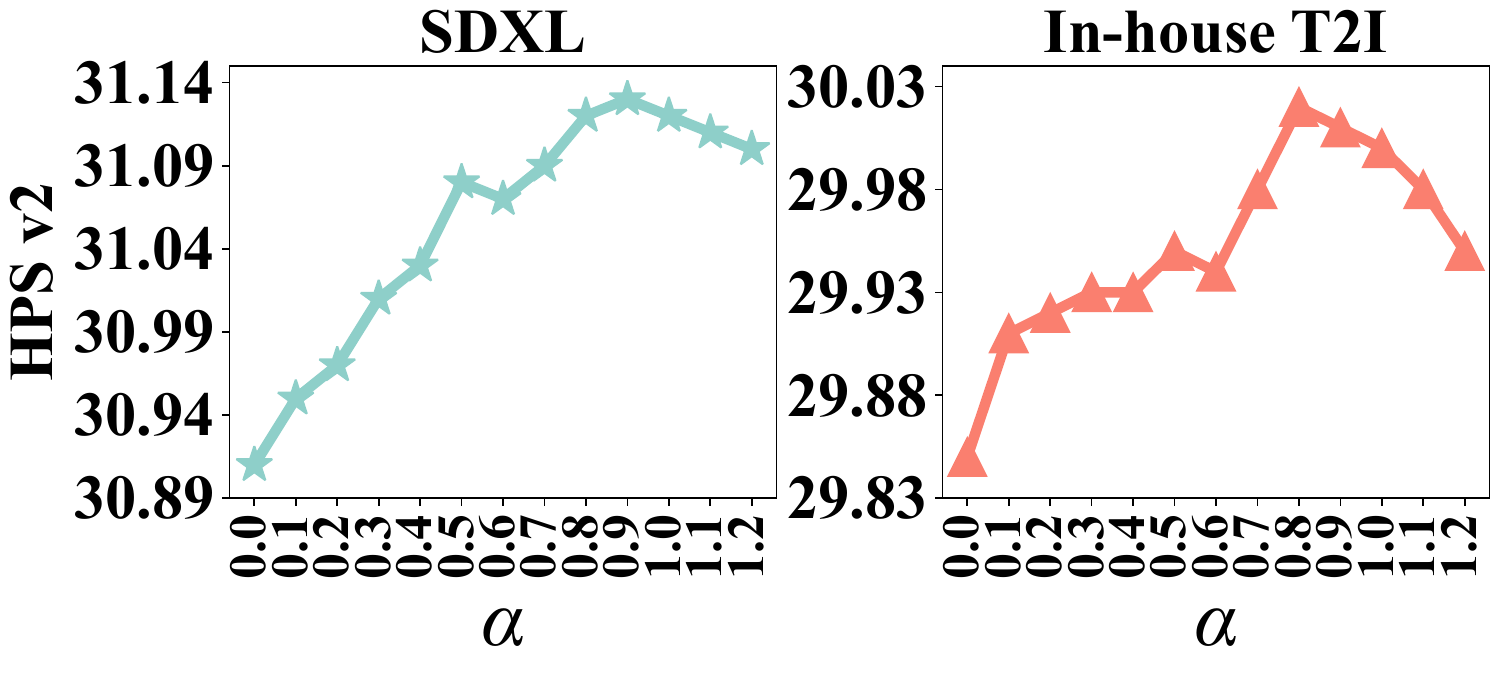} \label{fig:supp_alpha_paintings}}
    \hfill
    \subfigure[Photo]{\includegraphics[width=0.48\textwidth]{figs/alpha_photo.pdf} \label{fig:supp_alpha_photo}}
    \caption{Full ablation results on $\alpha$. We report the HPS v2 results on Anime, Concept-Art, Paintings and Photo.}
    \label{fig:supp_full_alpha}
\end{figure}

In Fig.~\ref{fig:supp_full_alpha}, we present the complete results of the ablation studies on $\alpha$, exploring the impact of varying its values. CHATS consistently outperforms CFG ($\alpha=0$) across almost all cases, as it better incorporates preference information during sampling. Any value of $\alpha$ within the range $[0.5, 0.7]$ performs well. Therefore, to demonstrate the generalizability of CHATS, we set $\alpha=0.5$ as the default choice for the main experiments, although $\alpha=0.7$ achieves the best performance.

\section{More Qualitative Results}

\begin{figure*}[t]
    \centering
    \includegraphics[width=.88\linewidth]{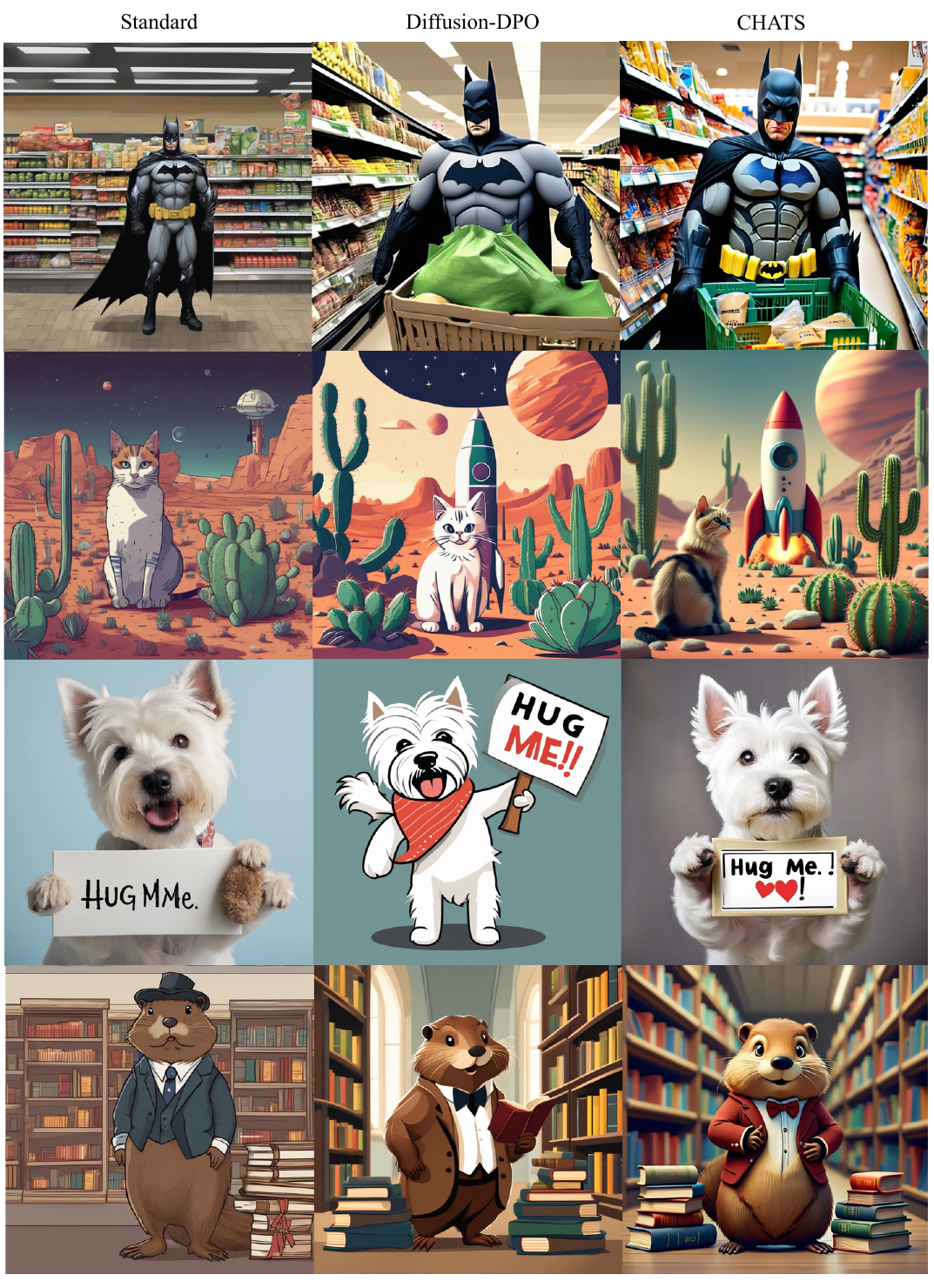}
    \caption{More qualitative comparisons. \textit{Prompts: 1) Batman is shown working as a Bagger at a grocery store. 2) A cat sitting besides a rocket on a planet with a lot of cactuses. 3) A West Highland white terrier holding a ``Hug me!'' sign. 4) 
    A beaver in formal attire stands beside books in a library.}}
    \label{fig:supp_qualitative_0}
\end{figure*}

\begin{figure*}[t]
    \centering
    \includegraphics[width=.88\linewidth]{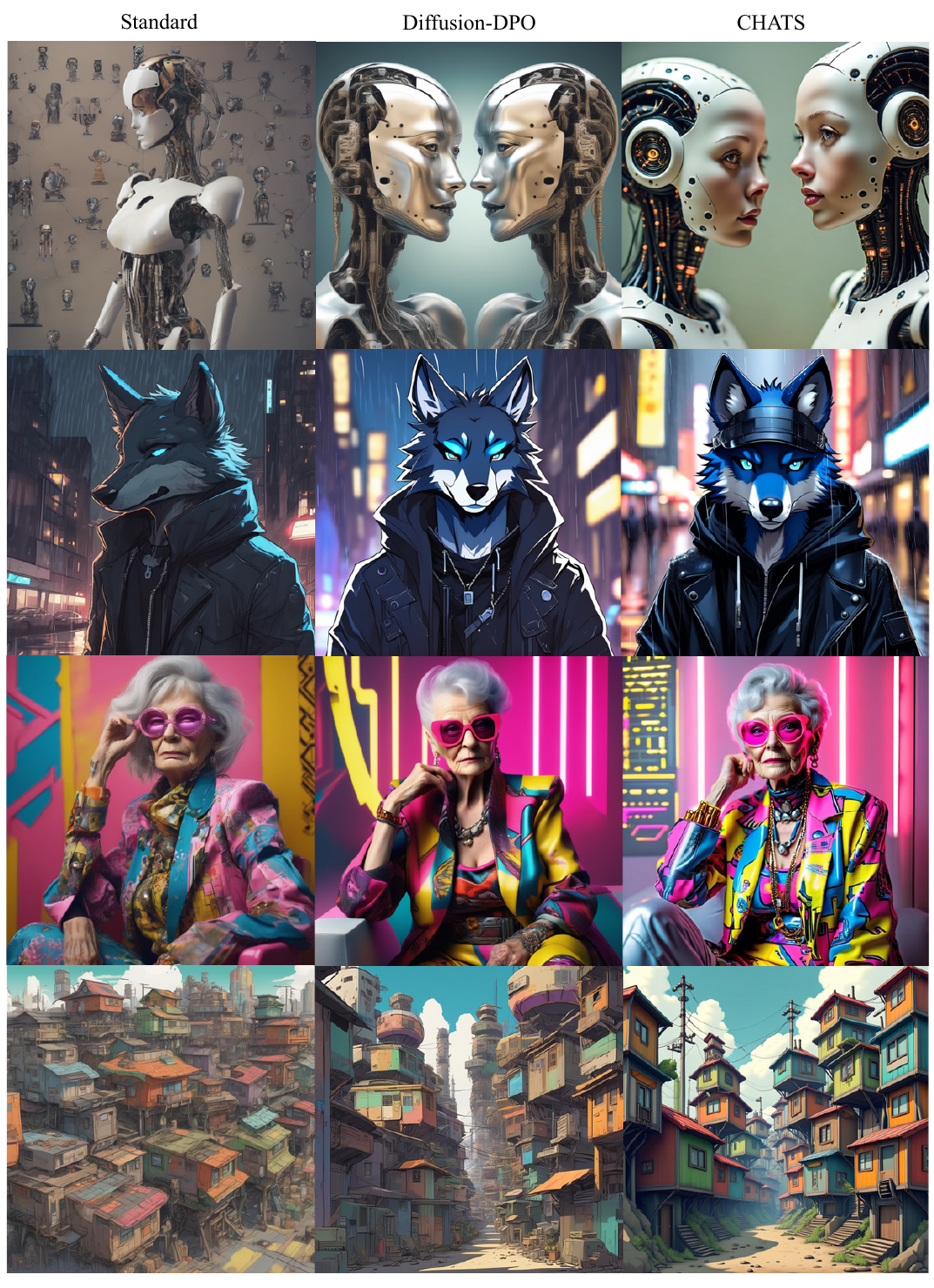}
    \caption{More qualitative comparisons. \textit{Prompts: 1) An anthropomorphic and surreal depiction of artificial intelligence's self-image. 2) Portrait of a male furry anthro Blue wolf fursona wearing black cyberpunk clothes in a city at night while it rains. 3) An elderly woman poses for a high fashion photoshoot in colorful, patterned clothes with a cyberpunk 2077 vibe. 4) Colorful scifi shanty town with metal rooftops and wooden and concrete walls in the style of Studio Ghibli and other anime influences.}}
    \label{fig:supp_qualitative_1}
\end{figure*}

\begin{figure*}[t]
    \centering
    \includegraphics[width=.88\linewidth]{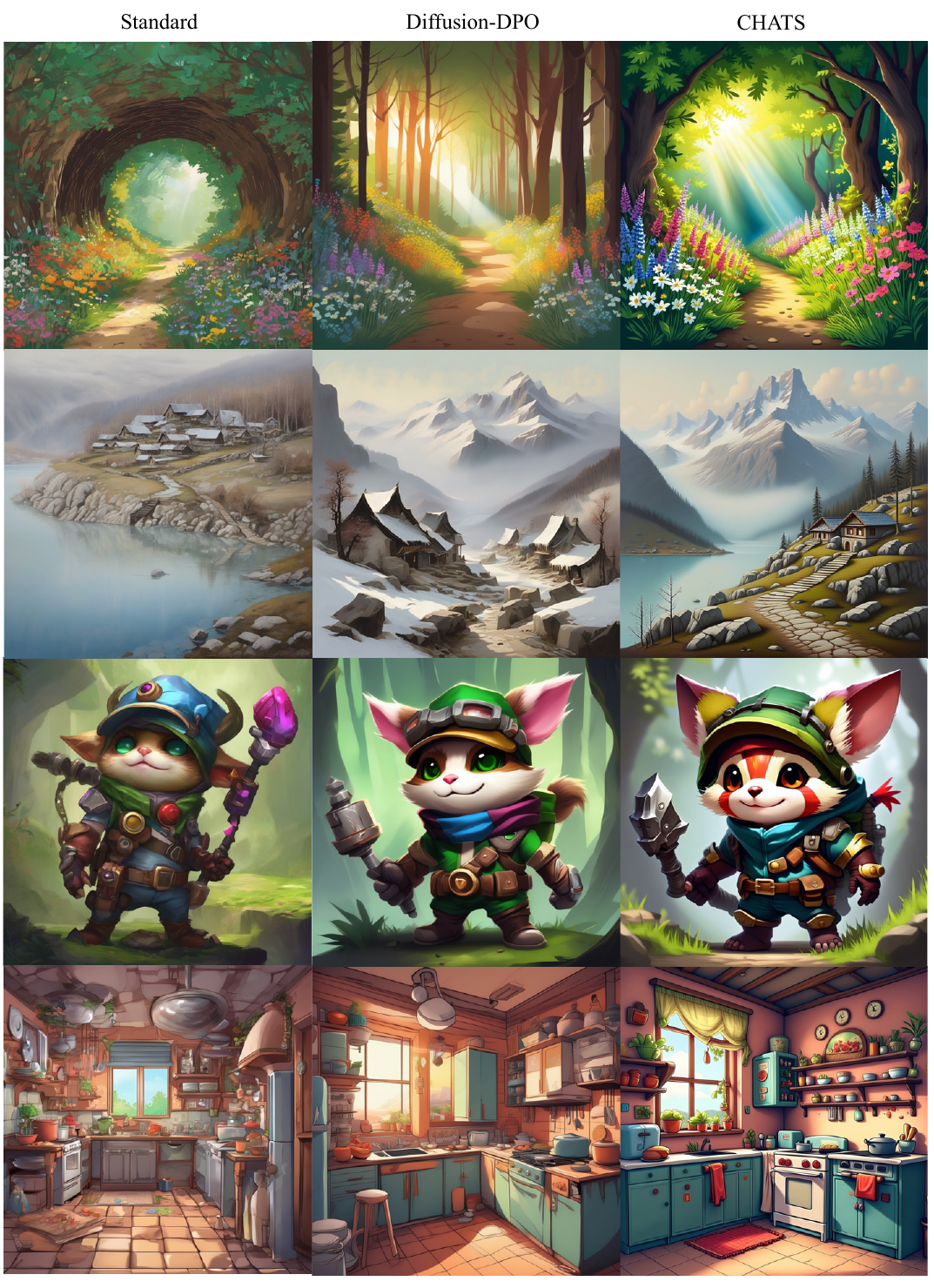}
    \caption{More qualitative comparisons. \textit{Prompts: 1) Colorful illustration of a forest tunnel illuminated by sunlight and filled with wildflowers. 2) A spring landscape painting featuring a treeless mountain village with melting lake ice, winding stone steps, and fog. 3) A digital painting of Teemo from League of Legends, wearing cyborg parts and a new skin, in a fantasy MMORPG style. 4) A digital painting of a fantasy kitchen environment with elements of cartoons, comics, and manga.}}
    \label{fig:supp_qualitative_2}
\end{figure*}

\begin{figure*}[t]
    \centering
    \includegraphics[width=.88\linewidth]{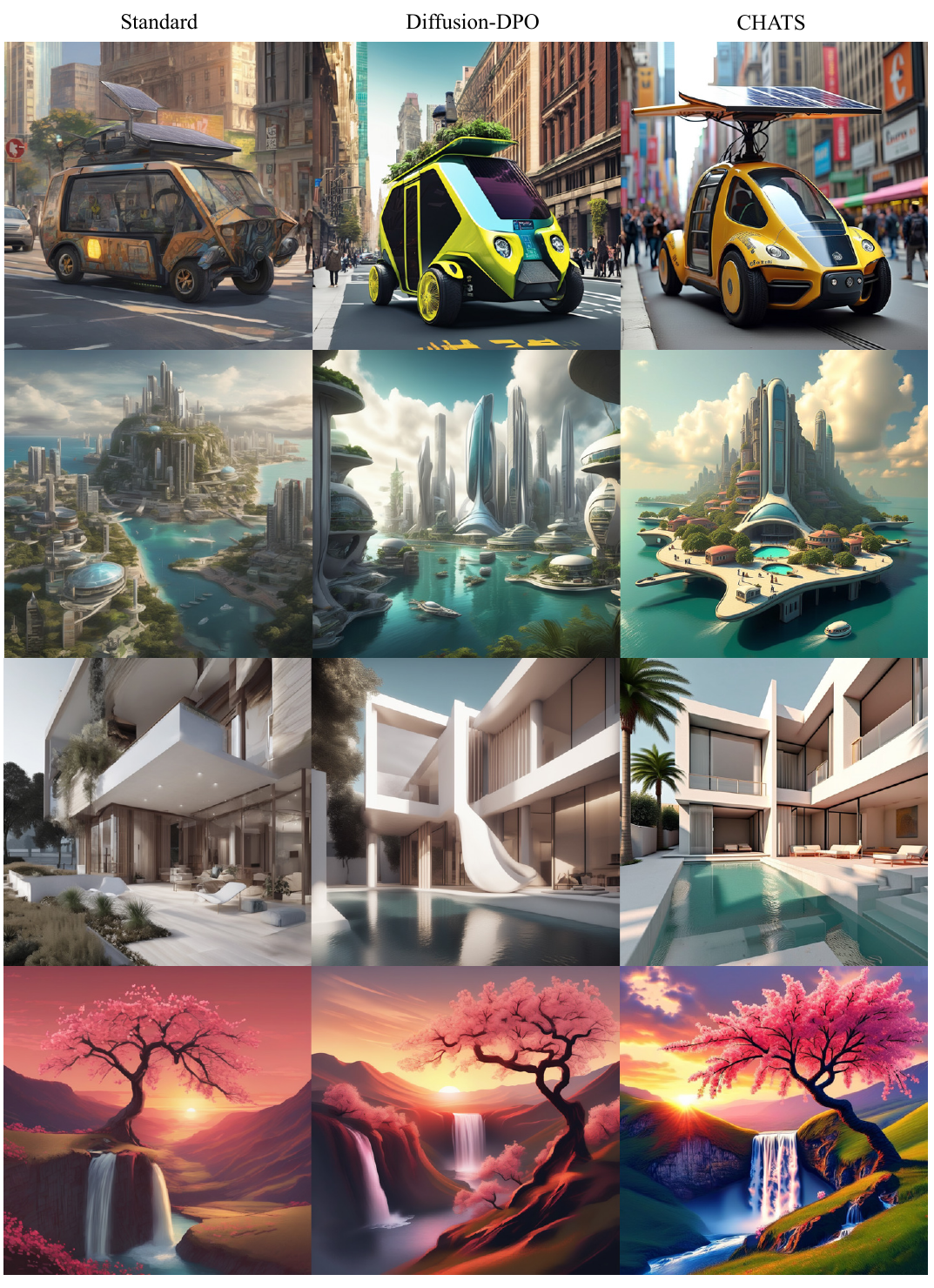}
    \caption{More qualitative comparisons. \textit{Prompts: 1) Solar punk vehicle in a bustling city. 2) A neofuturistic island city depicted in a photo-realistic illustration by five artists. 3) Architecture render with pleasing aesthetics. 4) A stylized digital art image of a cherry tree overlooking a valley with a waterfall during sunset.}}
    \label{fig:supp_qualitative_3}
\end{figure*}

\begin{figure*}[t]
    \centering
    \includegraphics[width=.88\linewidth]{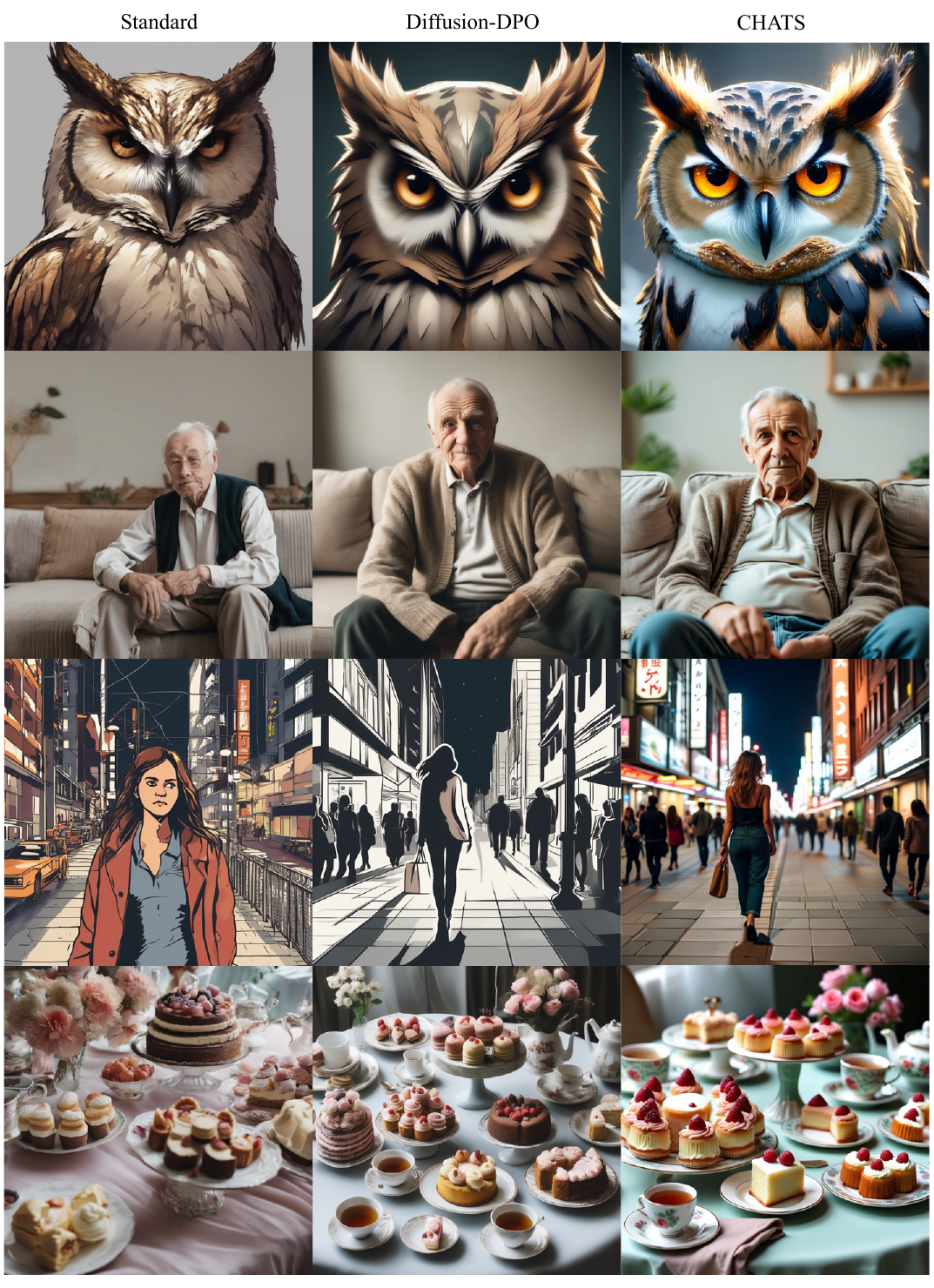}
    \caption{More qualitative comparisons. \textit{Prompts: 1) The image is a digital art headshot of an owlfolk character with high detail and dramatic lighting. 2) An elderly man is sitting on a couch. 3) Woman walking down the side walk of a busy night city. 4) A view of a table with a bunch of cakes and tea on it. }}
    \label{fig:supp_qualitative_4}
\end{figure*}

We present additional qualitative results in Fig.~\ref{fig:supp_qualitative_0} to~\ref{fig:supp_qualitative_4}. As shown in these figures, our CHATS significantly enhances the quality of generated images compared to the standard version of models. Furthermore, in comparison to Diffusion-DPO, our approach shows a superior capability in capturing and accurately rendering the intricate details described in the prompts.

\end{document}